\documentclass[10pt,twocolumn,letterpaper]{article}
\usepackage{titling}
\usepackage{iccv}
\usepackage{times}
\usepackage{epsfig}
\usepackage{graphicx}
\usepackage{amsmath}
\usepackage{amssymb}
\usepackage{algorithm}
\usepackage{algorithmic}
\usepackage{subcaption}
\usepackage{multirow}
\usepackage{booktabs}
\usepackage[title]{appendix}

\usepackage[pagebackref=true,breaklinks=true,letterpaper=true,colorlinks,bookmarks=false]{hyperref}

 \iccvfinalcopy 

\def\iccvPaperID{0000} 

\ificcvfinal\fi
\begin{document}

\title{Diverse Image Synthesis from Semantic Layouts via Conditional IMLE}

\author{Ke Li\thanks{Equal contribution.}\\
UC Berkeley\\
{\tt\small ke.li@eecs.berkeley.edu}
\and
Tianhao Zhang$^{\ast}$\\
Nanjing University\\
{\tt\small bryanzhang@smail.nju.edu.cn}
\and
Jitendra Malik\\
UC Berkeley\\
{\tt\small malik@eecs.berkeley.edu}
}

\maketitle

\begin{abstract}
Most existing methods for conditional image synthesis are only able to generate a single plausible image for any given input, or at best a fixed number of plausible images. In this paper, we focus on the problem of generating images from semantic segmentation maps and present a simple new method that can generate an arbitrary number of images with diverse appearance for the same semantic layout. Unlike most existing approaches which adopt the GAN~\cite{goodfellow2014generative,gutmann2014likelihood} framework, our method is based on the recently introduced Implicit Maximum Likelihood Estimation (IMLE)~\cite{li2018implicit} framework. Compared to the leading approach~\cite{chen2017photographic}, our method is able to generate more diverse images while producing fewer artifacts despite using the same architecture. The learned latent space also has sensible structure despite the lack of supervision that encourages such behaviour. Videos and code are available at \url{https://people.eecs.berkeley.edu/~ke.li/projects/imle/scene_layouts/}.

\end{abstract}

\begin{figure}[h]
    \centering
    \includegraphics[width=0.9\linewidth]{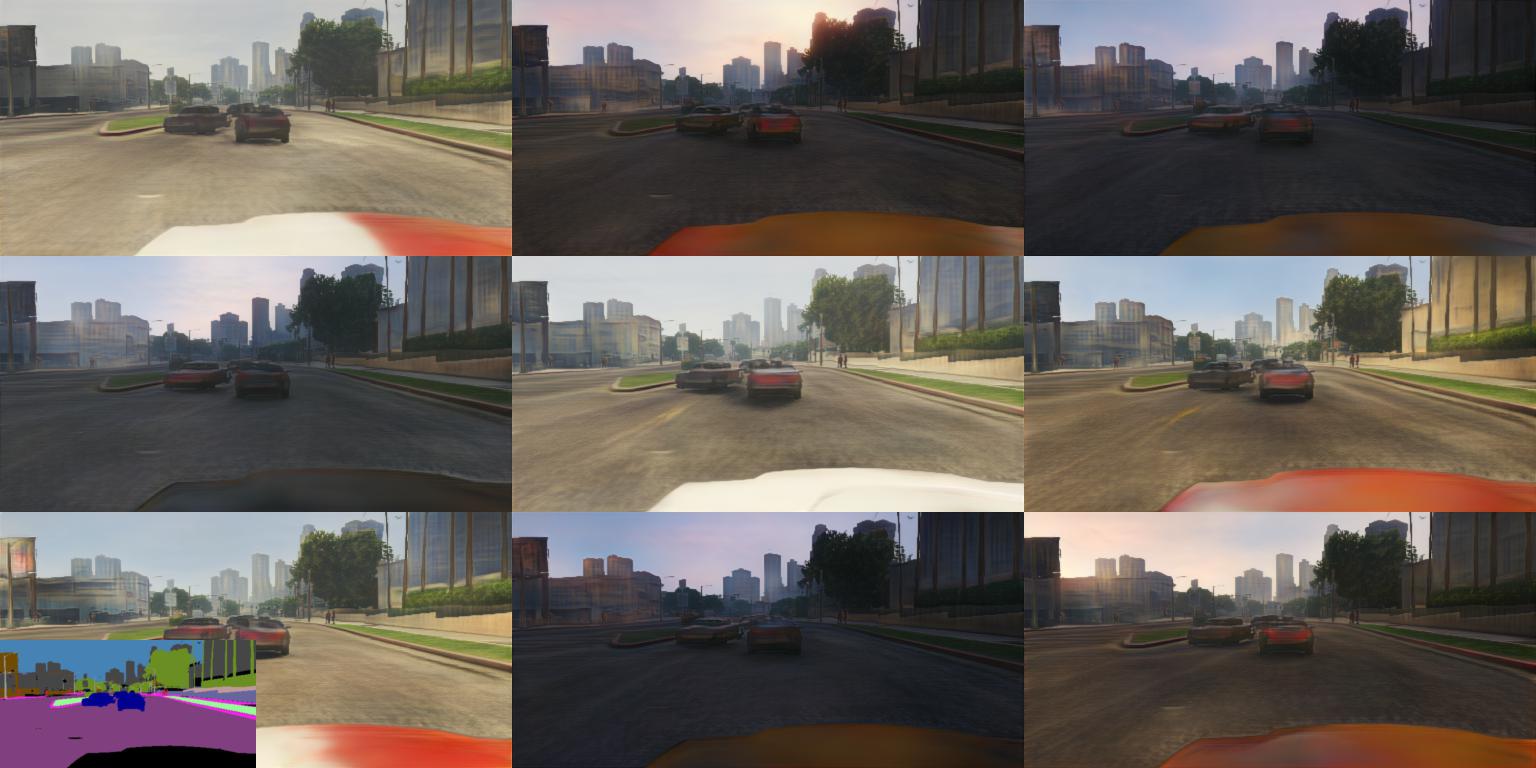}
    \caption{Samples generated by our model. The 9 images are samples generated by our model conditioned on the same semantic layout as shown at the bottom-left corner.}
    \label{fig:eg20}
\end{figure}
\section{Introduction}

Conditional image synthesis is a problem of great importance in computer vision. In recent years, the community has made great progress towards generating images of high visual fidelity on a variety of tasks. However, most proposed methods are only able to generate a single image given each input, even though most image synthesis problems are ill-posed, i.e.: there are multiple equally plausible images that are consistent with the same input. Ideally, we should aim to predict a distribution of all plausible images rather than just a single plausible image, which is a problem known as \emph{multimodal image synthesis}~\cite{zhu2017toward}. This problem is hard for two reasons:
\begin{enumerate}
    \item \emph{Model:} Most state-of-the-art approaches for image synthesis use generative adversarial nets (GANs)~\cite{goodfellow2014generative,gutmann2014likelihood}, which suffer from the well-documented issue of \emph{mode collapse}. In the context of conditional image synthesis, this leads to a model that generates only a single plausible image for each given input regardless of the latent noise and fails to learn the distribution of plausible images.  
    \item \emph{Data:} Multiple different ground truth images for the same input are not available in most datasets. Instead, only one ground truth image is given, and the model has to learn to generate other plausible images in an unsupervised fashion. 
\end{enumerate}

In this paper, we focus on the problem of multimodal image synthesis from semantic layouts, where the goal is to generate multiple \emph{diverse} images for the same semantic layout. Existing methods are either only able to generate a fixed number of images~\cite{chen2017photographic} or are difficult to train~\cite{zhu2017toward} due to the need to balance the training of several different neural nets that serve opposing roles. 

To sidestep these issues, unlike most image synthesis approaches, we step outside of the GAN framework and propose a method based on the recently introduced method of Implicit Maximum Likelihood Estimation (IMLE)~\cite{li2018implicit}. Unlike GANs, IMLE by design avoids mode collapse and is able to train the same types of neural net architectures as generators in GANs, namely neural nets with random noise drawn from an analytic distribution as input. 

This approach offers two advantages:
\begin{enumerate}
    \item Unlike \cite{chen2017photographic}, we can generate an arbitrary number of images for each input by simply sampling different noise vectors. 
    \item Unlike \cite{zhu2017toward}, which requires the simultaneous training of three neural nets that serve opposing roles, our model is much simpler: it only consists of a single neural net. Consequently, training is much more stable. 
\end{enumerate}

\section{Related Work}

\subsection{Unimodal Prediction}
Most modern image synthesis methods are based on generative adversarial nets (GANs)~\cite{goodfellow2014generative,gutmann2014likelihood}. Most of these methods are capable of producing only a single image for each given input, due to the problem of mode collapse. Various work has explored conditioning on different types of information. Various methods condition on a scalar that only contains little information, such as object category and attribute~\cite{mirza2014conditional,gauthier2014conditional,denton2015deep}. Other methods condition on richer labels, such as text description~\cite{reed2016learning}, surface normal maps~\cite{wang2016generative}, previous frames in a video~\cite{mathieu2015deep,vondrick2016generating} and images~\cite{yoo2016pixel,isola2017image,zhu2017unpaired}. Some methods only condition on inputs images in the generator, but not in the discriminator~\cite{pathak2016context,ledig2017photo,zhu2016generative,li2016precomputed}. \cite{kaneko2017generative,reed2016learning,sangkloy2017scribbler} explore conditioning on attributes that can be modified manually by the user at test time; these methods are not true multimodal methods because they require manual changes to the input (rather than just sampling from a fixed distribution) to generate a different image. 

Another common approach to image synthesis is to treat it as a simple regression problem. To ensure high perceptual quality, the loss is usually defined on some transformation of the raw pixels. This paradigm has been applied to super-resolution \cite{bruna2015super,johnson2016perceptual}, style transfer \cite{johnson2016perceptual} and video frame prediction \cite{srivastava2015unsupervised,oh2015action,finn2016unsupervised}. These methods are by design unimodal methods because neural nets are functions, and so can only produce point estimates. 

Various methods have been developed for the problem of image synthesis from semantic layouts. For example, Karacan \etal \cite{karacan2016learning} developed a conditional GAN-based model for generating images from semantic layouts and labelled image attributes. It is important to note that the method requires supervision on the image attributes and is therefore a unimodal method. Isola \etal~\cite{isola2017image} developed a conditional GAN that can generate images solely from semantic layout. However, it is only able to generate a single plausible image for each semantic layout, due to the problem of mode collapse in GANs. Wang \etal~\cite{wang2017high} further refined the approach of \cite{isola2017image}, focusing on the high-resolution setting. While these methods are able to generate images of high visual fidelity, they are all unimodal methods. 

\subsection{Fixed Number of Modes}

A simple approach to generate a fixed number of different outputs for the same input is to use different branches or models for each desired output. For example, \cite{guzman2012multiple} proposed a model that outputs a fixed number of different predictions simultaneously, which was an approach adopted by Chen and Koltun \cite{chen2017photographic} to generate different images for the same semantic layout. Unlike most approaches, \cite{chen2017photographic} did not use the GAN framework; instead it uses a simple feedforward convolutional network. On the other hand, Ghosh \etal~\cite{ghosh2017multi} uses a GAN framework, where multiple generators are introduced, each of which generates a different mode. The above methods all have two limitations: (1) they are only able to generate a fixed number of images for the same input, and (2) they cannot generate continuous changes.  

\subsection{Arbitrary Number of Modes}

A number of GAN-based approaches propose adding learned regularizers that discourage mode collapse. BiGAN/ALI~\cite{donahue2016adversarial,dumoulin2016adversarially} trains a model to reconstruct the latent code from the image; however, when applied to the conditional setting, significant mode collapse still occurs because the encoder is not trained until optimality and so cannot perfectly invert the generator. VAE-GAN~\cite{larsen2015autoencoding} combines a GAN with a VAE, which does not suffer from mode collapse. However, image quality suffers because the generator is trained on latent code sampled from the encoder/approximate posterior, and is never trained on latent code sampled from the prior. At test time, only the prior is available, resulting in a mismatch between training and test conditions. Zhu \etal~\cite{zhu2017toward} proposed Bicycle-GAN, which combines both of the above approaches. While this alleviates the above issues, it is difficult to train, because it requires training three different neural nets simultaneously, namely the generator, the discriminator and the encoder. Because they serve opposing roles and effectively regularize one another, it is important to strike just the right balance, which makes it hard to train successfully in practice. 

A number of methods for colourization~\cite{charpiat2008automatic,zhang2016colorful,larsson2016learning} predict a discretized marginal distribution over colours of each individual pixel. While this approach is able to capture multimodality in the marginal distribution, ensuring global consistency between different parts of the image is not easy, since there are correlations between the colours of different pixels. This approach is not able to learn such correlations because it does not learn the joint distribution over the colours of \emph{all} pixels. 
\begin{figure*}
    \centering
    \begin{subfigure}[t]{0.309\textwidth}
        \centering
        \includegraphics[width=\textwidth]{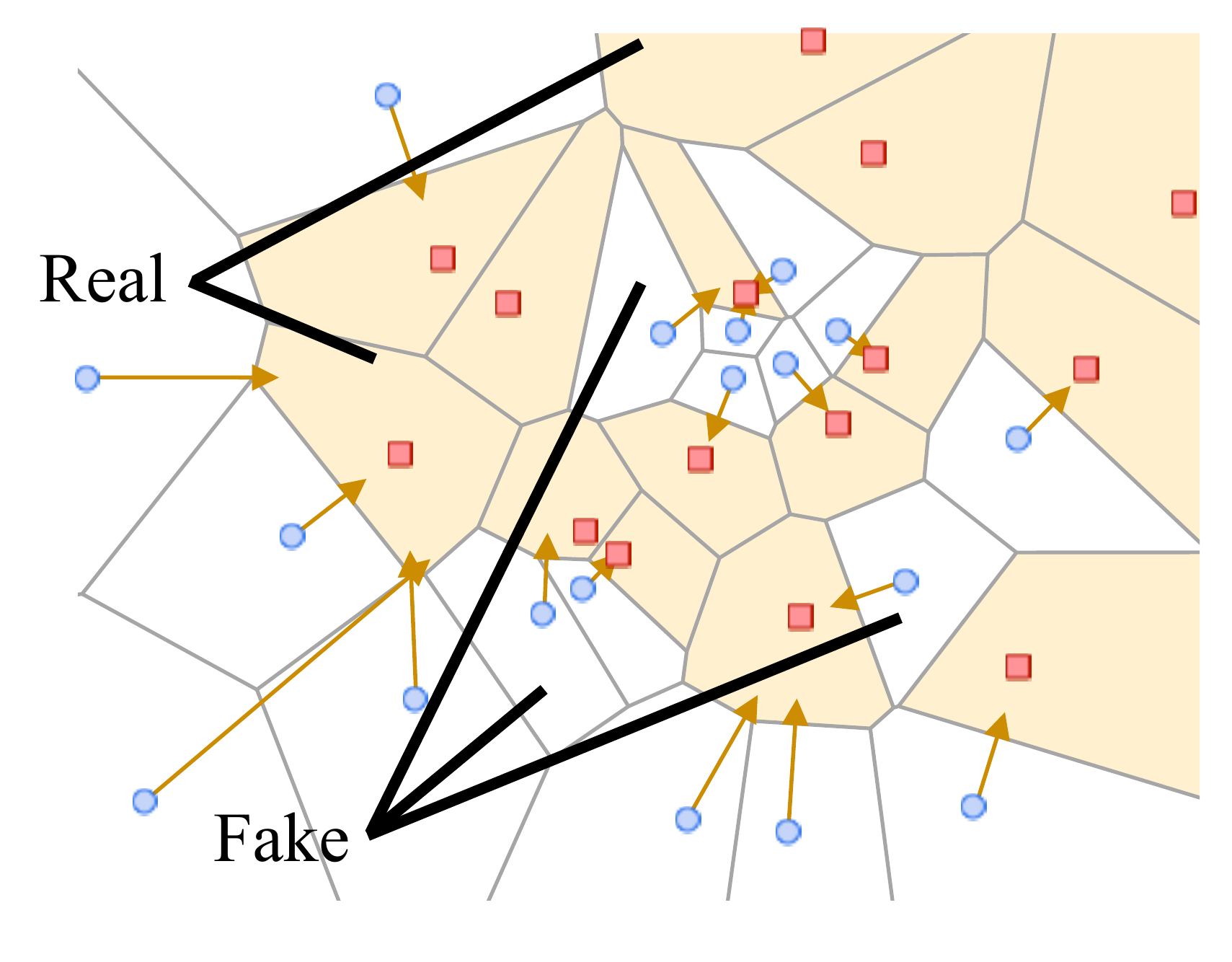}
        \caption{GAN\protect\\(Step 1)}\label{fig:schematic_gan1}
    \end{subfigure}
    \begin{subfigure}[t]{0.373\textwidth}
        \centering
        \includegraphics[width=\textwidth]{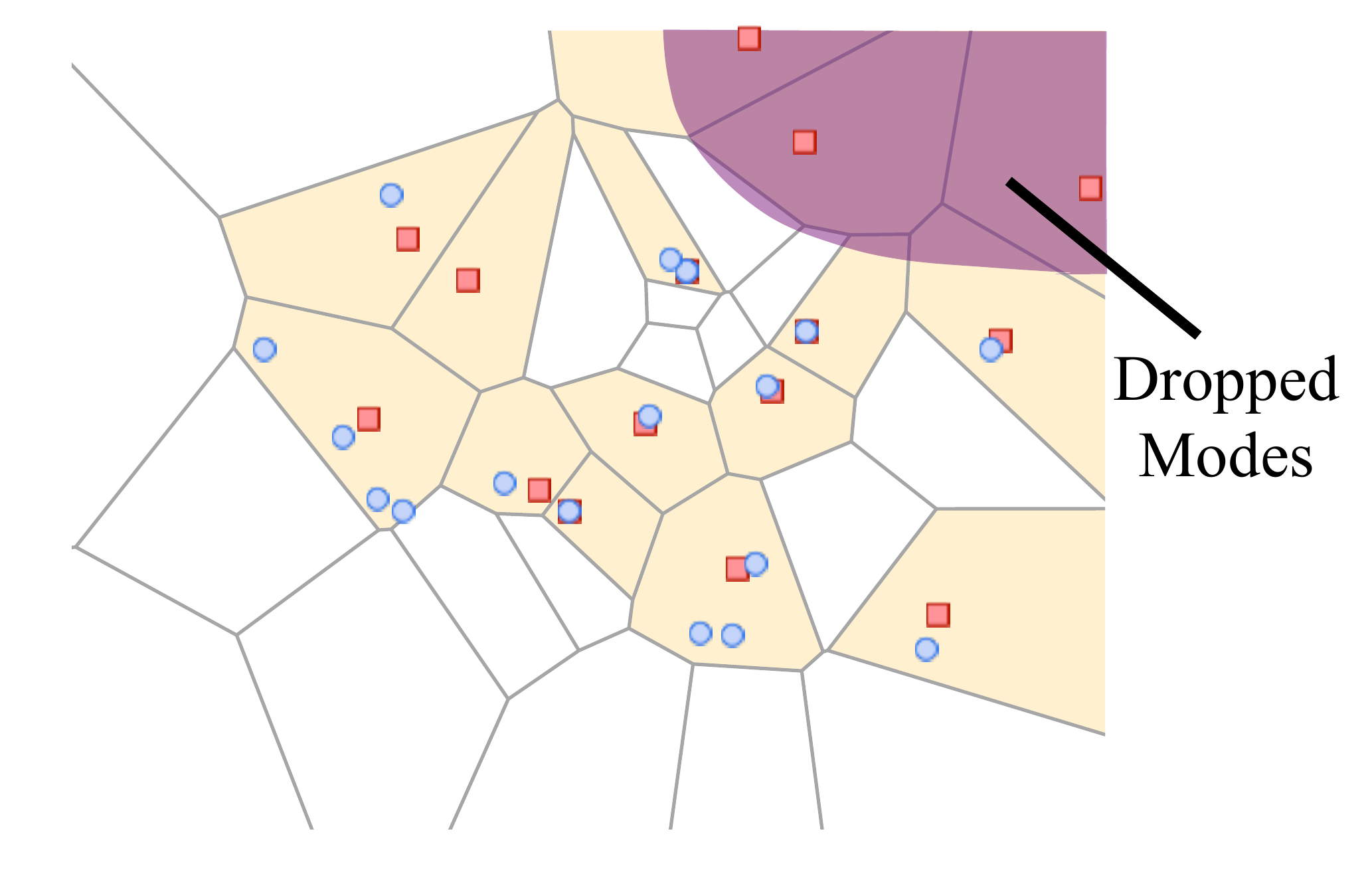}
        \caption{GAN\protect\\(Step 2)}\label{fig:schematic_gan2}
    \end{subfigure}
    \begin{subfigure}[t]{0.298\textwidth}
        \centering
        \includegraphics[width=\textwidth]{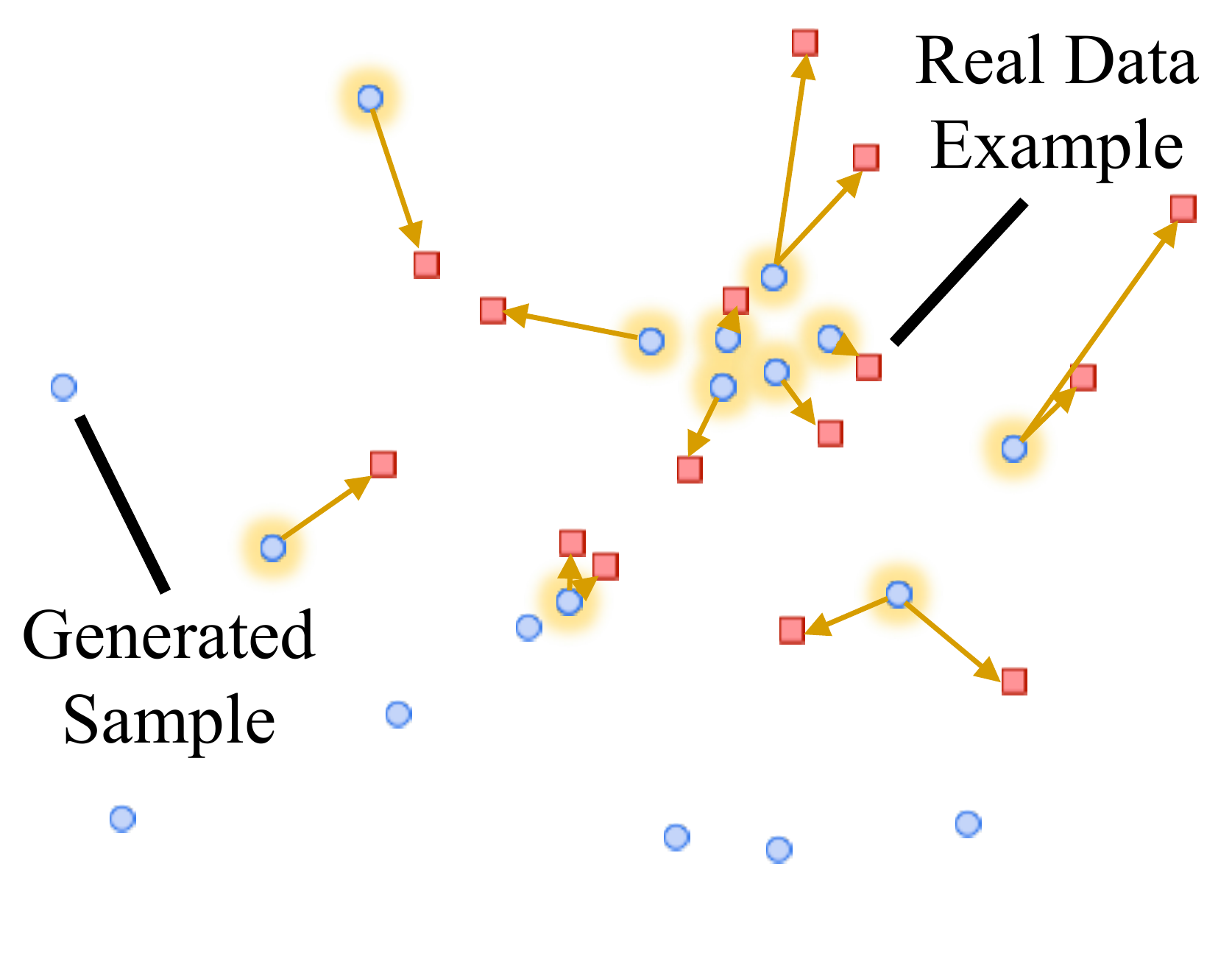}
        \caption{IMLE}\label{fig:schematic_imle}
    \end{subfigure}
    \caption{\label{fig:schematic}(a-b) How a (unconditional) GAN collapses modes (here we show a GAN with 1-nearest neighbour discriminator for simplicity). The blue circles represent generated images and the red squares represent real images. The yellow regions represent those classified as real by the discriminator, whereas the white regions represent those classified as fake. As shown, when training the generator, each \emph{generated image} is essentially pushed towards the nearest \emph{real image}. Some real images may not be selected by any generated image during training and therefore could be ignored by the trained generator -- this is a manifestation of mode collapse. (c) An illustration of how Implicit Maximum Likelihood Estimation (IMLE) works. IMLE avoids mode collapse by reversing the direction in which generated images are matched to real images. Instead of pushing each generated image towards the nearest real image, for every real image, it pulls the nearest generated image towards it -- this ensures that all real images are matched to some generated image, and no real images are ignored.}
    
\end{figure*}

\section{Method}

Most state-of-the-art approaches for conditional synthesis rely on the conditional GAN framework. Unfortunately, GANs suffer from the well-known problem of mode collapse, and in the context of conditional image synthesis, this causes the generator to ignore the latent input noise vector and always generate the same output image for the same input label, regardless of what the value of the latent noise vector. So, to generate different output images for the same input label, we must solve the underlying problem of mode collapse. 

\subsection{Why Mode Collapse Happens}

We first consider the unconditional setting, where there is no input label. As shown in Figure~\ref{fig:schematic}(a-b), in a GAN, each generated image is made similar to \emph{some} real image. Some images may not be selected by any real image. So after training, the generator will not be able to generate any image that is similar to the unselected real images, so it effectively ignores these images. In the language of probabilistic modelling, real images can be viewed as samples from some underlying true distribution of natural images, and the generator ignoring some of the real images means that the generative model assigns low probability density to these images. So, the modes (i.e.: the local maxima in the probability density) of the true distribution of natural images that represent the ignored images are not modelled by the generator; hence the name ``mode collapse''. In the conditional setting, typically only one ground truth output image is available for every input label. As a result, mode collapse becomes more problematic, because the conditional distribution modelled by the generator will collapse to a single mode around the sole ground truth output image. This means that the generator will not be able to output any other equally plausible output image. 

\subsection{IMLE}

The method of Implicit Maximum Likelihood Estimation (IMLE)~\cite{li2018implicit} solves mode collapse by reversing the direction in which generated images are matched to real images. In a GAN, each generated image is effectively matched to a real image. In IMLE, each real image is matched to a generated image. This ensures that all real images are matched, and no real images are left out. As shown in Figure~\ref{fig:schematic}(c), IMLE then tries to make each matched generated image similar to the real images they are matched to. Mathematically, it solves the optimization problem below. Here, $\mathbf{z}_j$'s denote randomly sampled latent input noise vectors, $\mathbf{y}_i$'s denote ground truth images, and $T_\theta$ denotes a neural net whose architecture is the same as the generator in GANs. 
\begin{equation*}
    \min_{\theta}\mathbb{E}_{\mathbf{z}_1,\ldots,\mathbf{z}_m} \left[\frac{1}{n}\sum_{i=1}^n\min_{j=1,\dots,m}||T_\theta(\mathbf{z}_j)-\mathbf{y}_i||_2^2\right]
\end{equation*}

\subsection{Conditional IMLE}

In the conditional setting, the goal is to model a family of conditional distributions, each of is conditioned on a different input label, i.e.: $\left\{p(\mathbf{y}|\mathbf{x} = \mathbf{x}_i)\right\}_{i=1,\ldots,n}$, where $\mathbf{x}_i$'s denote ground truth input images, and $\mathbf{y}$ denotes the generated output image. So, conditional IMLE~\cite{li2018super} differs from standard IMLE in two ways: first, the input label is passed into the neural net $T_\theta$ in addition to the latent input noise vector, and second, a ground truth output image can only be matched to an output image generated from its corresponding ground truth input label (i.e.: output images generated from an input label that is different from the current ground truth input label cannot be matched to the current ground truth output image). Concretely, conditional IMLE solves the following optimization problem, where $\mathbf{z}_{i,j}$'s denote randomly sampled latent input noise vectors and $\mathbf{y}_i$'s denote ground truth images:
\begin{equation*}
    \min_{\theta}\mathbb{E}_{\mathbf{z}_{1,1},\ldots,\mathbf{z}_{n,m}} \left[\frac{1}{n}\sum_{i=1}^n\min_{j=1,\dots,m}||T_\theta(\mathbf{x}_i, \mathbf{z}_{i,j})-\mathbf{y}_i||_2^2\right]
\end{equation*}

\subsection{Probabilistic Interpretation}

Image synthesis can be viewed as a probabilistic modelling problem. In unconditional image synthesis, the goal is the model the marginal distribution over images, i.e.: $p(\mathbf{y})$, whereas in conditional image synthesis, the goal is to model the conditional distribution 
$p(\mathbf{y}|\mathbf{x})$. In a conditional GAN, the probabilistic model is chosen to be an \emph{implicit probabilistic model}. Unlike classical (also known as prescribed) probabilistic models like CRFs, implicit probabilistic models are not defined by a formula for the probability density, but rather a procedure for drawing samples from them. The probability distributions they define are the distributions over the samples, so even though the formula for the probability density of these distributions may not be in closed form, the distributions themselves are valid and well-defined. The generator in GANs is an example of an implicit probabilistic model. It is defined by the following sampling procedure, where $T_\theta$ is a neural net: 

\begin{enumerate}
    \item Draw $\mathbf{z}\sim \mathcal{N}(\mathbf{0},\mathbf{I})$
    \item Return $\mathbf{y} := T_\theta(\mathbf{x},\mathbf{z})$ as a sample
\end{enumerate}

In classical probabilistic models, learning, or in other words, parameter estimation, is performed by maximizing the log-likelihood of the ground truth images, either exactly or approximately. This is known as maximum likelihood estimation (MLE). However, in implicit probabilistic models, this is in general not feasible: because the formula for the probability density may not be in closed form, the log-likelihood function, which the sum of the log-densities of the model evaluated at each ground truth image, cannot be in general written down in closed form. The GAN can be viewed as an alternative way to estimate the parameters of the probabilistic model, but it has one critical issue of mode collapse. As a result, the learned model distribution could capture much less variation than what the data exhibits. On the other hand, MLE never suffers from this issue: because mode collapse entails assigning very low probability density to some ground truth images, this would make the likelihood very low, because likelihood is the product of the densities evaluated at each ground truth image. So, maximizing likelihood will never lead to mode collapse. This implies that GANs cannot approximate maximum likelihood, and so the question is: is there some other algorithm that can? IMLE was designed with goal in mind, and can be shown to maximize a lower bound on the log-likelihood under mild conditions. Like GANs, IMLE does not need the formula for the probability density of the model to be known; unlike GANs, IMLE approximately maximizes likelihood, and so cannot collapse modes. Another added advantage comes from the fact that IMLE does not need a discriminator nor adversarial training. As a result, training is much more stable -- there is no need to balance the capacity of the generator with that of the discriminator, and so much less hyperparameter tuning is required. 

\subsection{Formulation}

For the task of image synthesis from semantic layouts, we take $\mathbf{x}$ to be the input semantic segmentation map and $\mathbf{y}$ to be the generated output image. (Details on the representation of $\mathbf{x}$ are in the supplementary material.) The conditional probability distribution that would like to learn is $p(\mathbf{y}|\mathbf{x})$. A plausible image that is consistent with the input segmentation $\mathbf{x}$ is a mode of this distribution; because there could be many plausible images that are consistent with the same segmentation, $p(\mathbf{y}|\mathbf{x})$ usually has multiple modes (and is therefore multimodal). A method that performs unimodal prediction can be seen as producing a point estimate of this distribution. To generate multiple plausible images, a point estimate is not enough; instead, we need to estimate the full distribution. 

We generalize conditional IMLE by using a different distance metric $\mathcal{L}(\cdot,\cdot)$, namely a perceptual loss based on VGG-19 features~\cite{simonyan2014very}, the details of which are in the supplementary material. The modified algorithm is presented in Algorithm~\ref{alg:A}. 

\begin{algorithm}\caption{\label{alg:A}Conditional IMLE}
\begin{algorithmic}
\STATE \textbf{Input} Training semantic segmentation maps $\{\mathbf{x}_i\}_{i=1}^n$ and the corresponding ground truth images $\{\mathbf{y}_i\}_{i=1}^n$
\STATE\textbf{Initialize} parameters $\theta$ for neural net $T_\theta$
\FOR{epoch = $1$ \textbf{to} $E$}
\STATE Pick a random batch $S\subseteq \{1,\ldots,n\}$ 
\FOR{$i \in S$}
\STATE Generate $m$ i.i.d random vectors $\{\mathbf{z}_{i,1},\ldots,\mathbf{z}_{i,m}\}$
\FOR{$j = 1\ \textbf{to}\ m$}
\STATE $\widetilde{\mathbf{y}}_{i,j}\gets T_\theta(\mathbf{x}_i,\mathbf{z}_{i,j})$
\ENDFOR
\STATE $\sigma(i)\gets \arg\min_{j\in\{1,\ldots,m\}}\mathcal{L}(\mathbf{y}_i,\widetilde{\mathbf{y}}_{i,j})$
\ENDFOR
\FOR{$k$ = 1 \textbf{to} $K$}
\STATE Pick a random batch $\widetilde{S}\subseteq S$
\STATE $\theta\gets \theta - \eta\nabla_\theta\left(\sum_{i\in \widetilde{S}}\mathcal{L}(\mathbf{y}_i,\widetilde{\mathbf{y}}_{i,\sigma(i)})\right) / |\widetilde{S}|$
\ENDFOR
\ENDFOR
\end{algorithmic}
\end{algorithm}
\subsection{Architecture}\label{sec:arch}

To allow for direct comparability to Cascaded Refinement Networks (CRN)~\cite{chen2017photographic}, which is the leading method for multimodal image synthesis from semantic layouts, we use the same architecture as CRN, with minor modifications to convert CRN into an implicit probabilistic model. 

The vanilla CRN synthesizes only one image for the same semantic layout input. To model an arbitrary number of modes, we add additional input channels to the architecture and feed random noise $\mathbf{z}$ via these channels. Because the noise is random, the neural net can now be viewed as a (implicit) probabilistic model. 

\paragraph{Noise Encoder} Because the input segmentation maps are provided at high resolutions, the noise vector $\mathbf{z}$, which is concatenated to the input channel-wise, could be very high-dimensional, which could hurt sample efficiency and therefore training speed. To solve this, we propose forcing the noise to lie on a low-dimensional manifold. To this end, we add a noise encoder module, which is a 3-layer convolutional net that takes the segmentation $\mathbf{x}$ and a lower-dimensional noise vector $\widetilde{\mathbf{z}}$ as input and outputs a noise vector $\mathbf{z}'$ of the same size as $\mathbf{z}$. We replace $\mathbf{z}$ with $\mathbf{z}'$ and leave the rest of the architecture unchanged.

\subsection{Dataset and Loss Rebalancing}
In practice, we found datasets can be strongly biased towards objects with relatively common appearance. As a result, na\"{i}ve training can result in limited diversity among the images generated by the trained model. To address this, we propose two strategies to rebalance the dataset and loss, the details of which are in the supplementary material. 
\section{Experiment}
\begin{figure}
    \centering
    \includegraphics[width=0.9\linewidth]{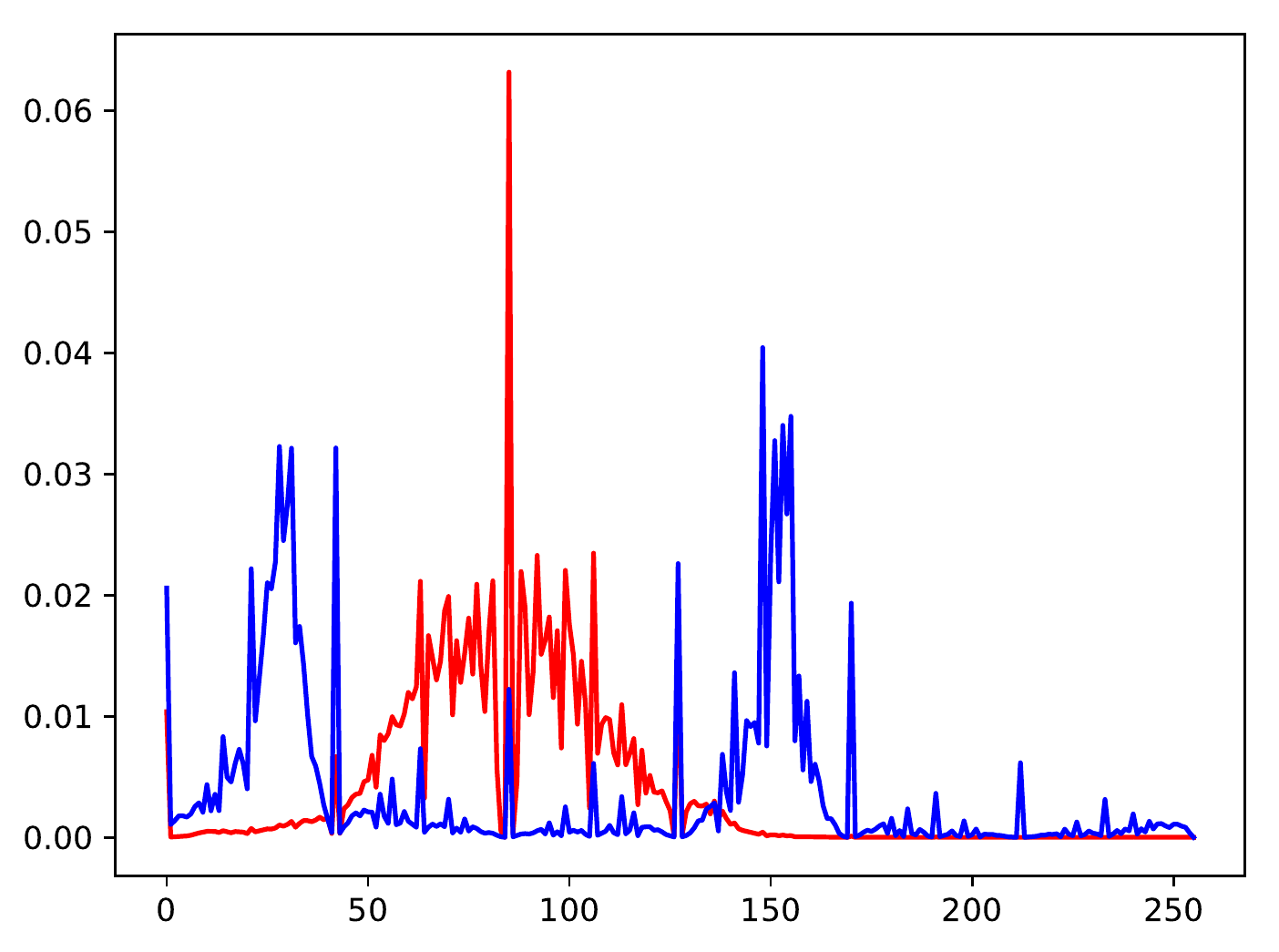}
    \caption{Comparison of histogram of hues between two datasets. Red is Cityscapes and blue is GTA-5. }
    \label{fig:hist}
\end{figure}
\begin{figure*}
    \centering
    \begin{subfigure}{0.47\textwidth}
    \includegraphics[width=\linewidth]{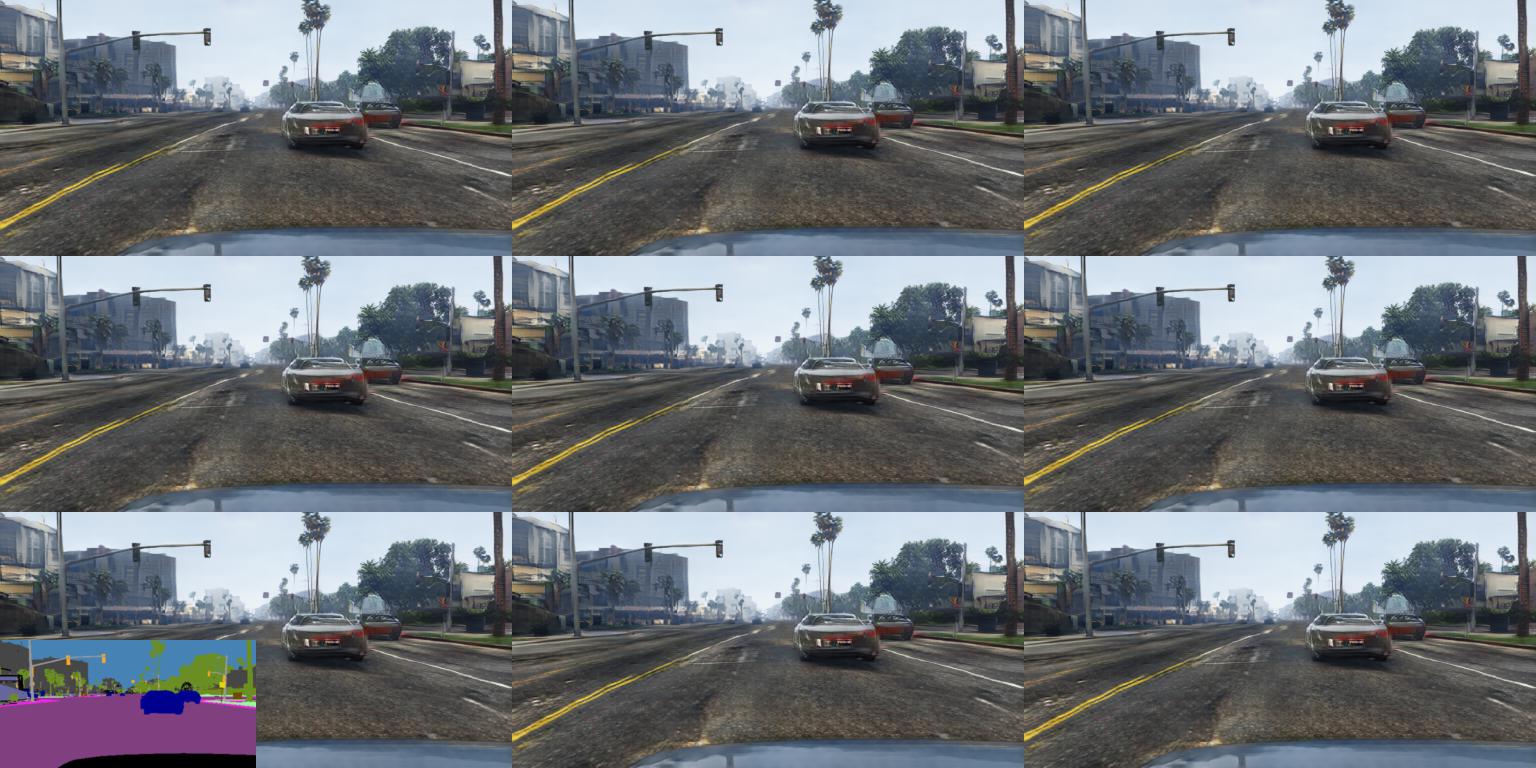}
    \caption{Pix2pix-HD+noise}
    \end{subfigure}
    \begin{subfigure}{0.47\textwidth}
    \includegraphics[width=\linewidth]{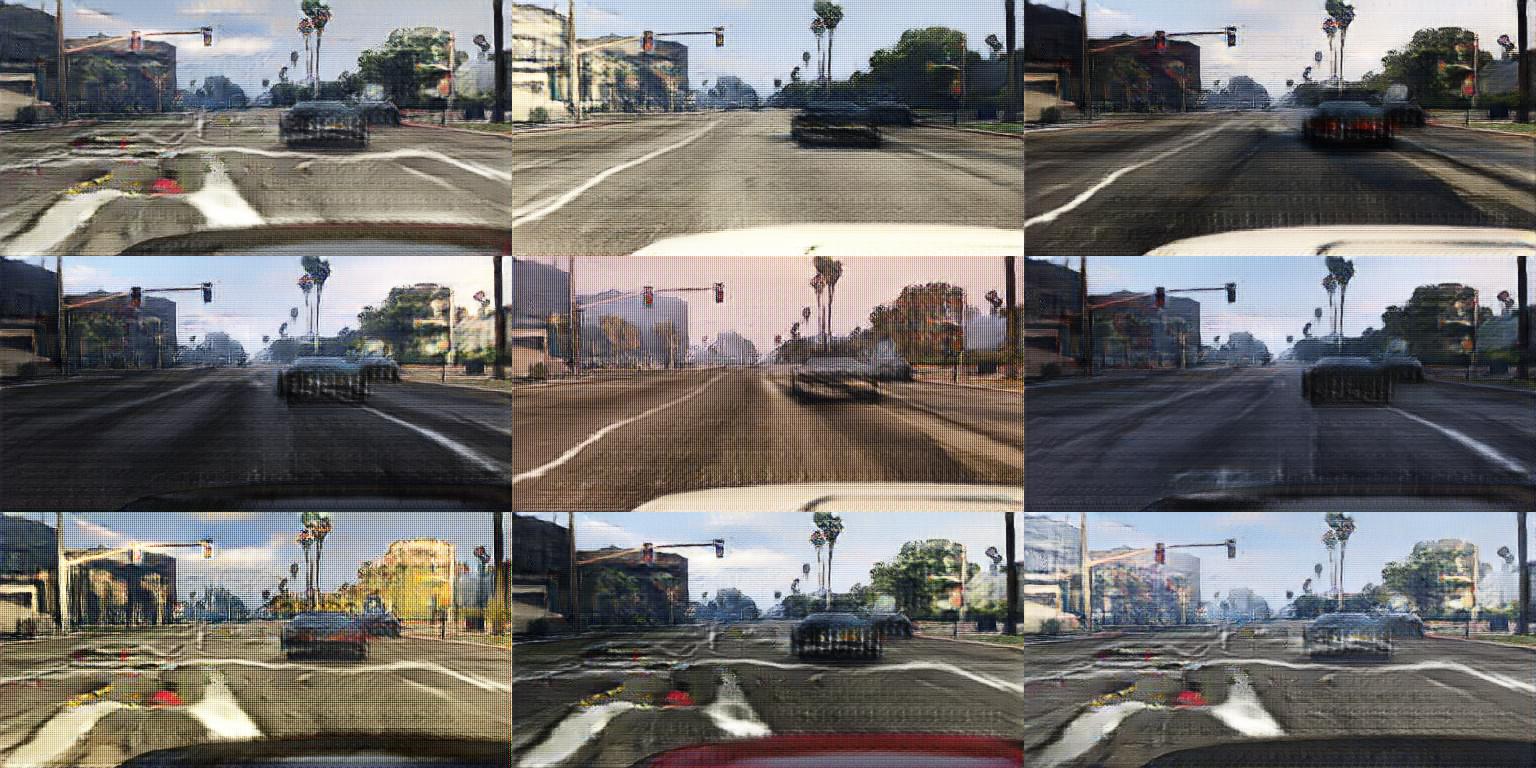}
    \caption{BicycleGAN}
    \end{subfigure}
    \begin{subfigure}{0.47\textwidth}
    \includegraphics[width=\linewidth]{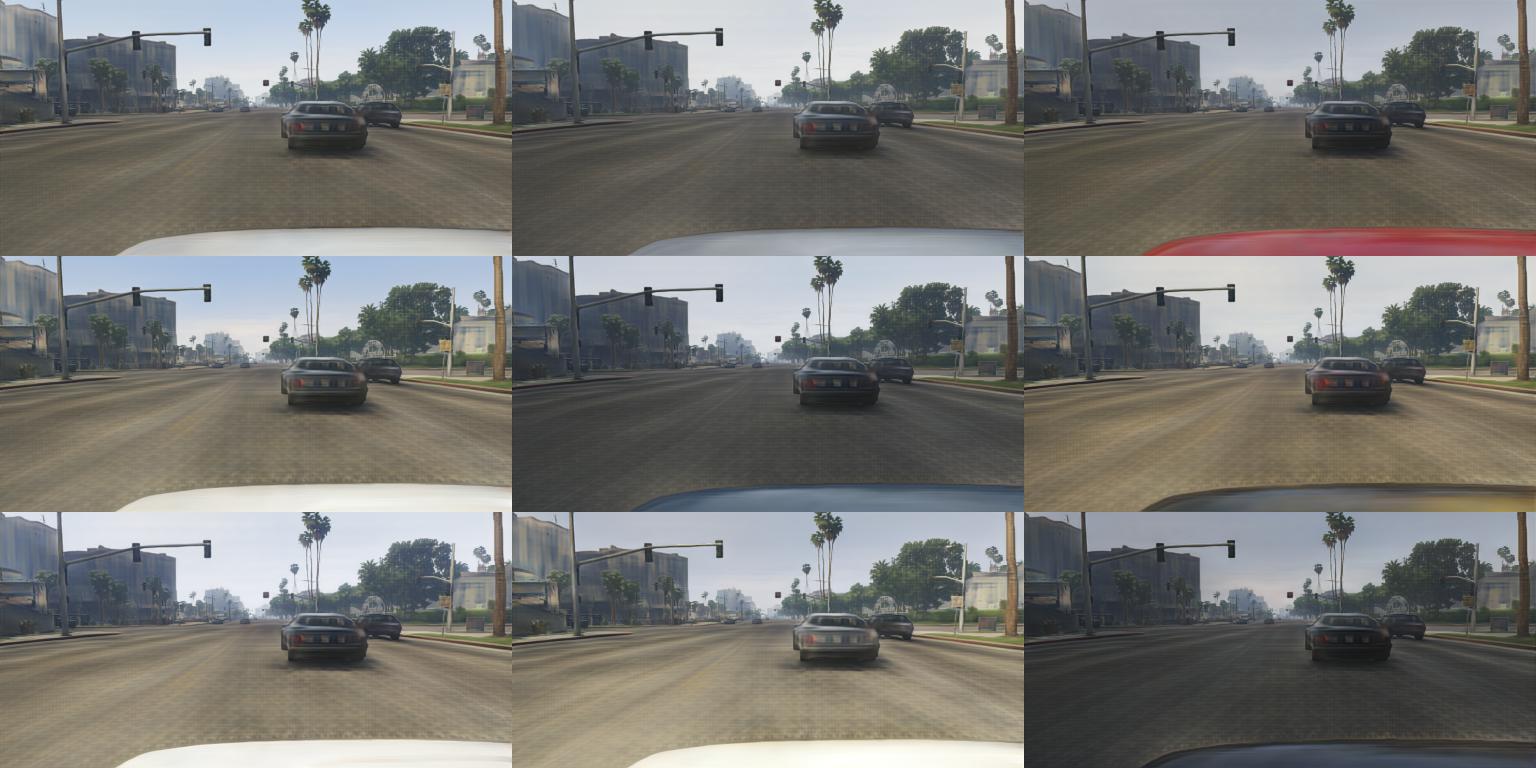}
    \caption{CRN}
    \end{subfigure}
    \begin{subfigure}{0.47\textwidth}
    \includegraphics[width=\linewidth]{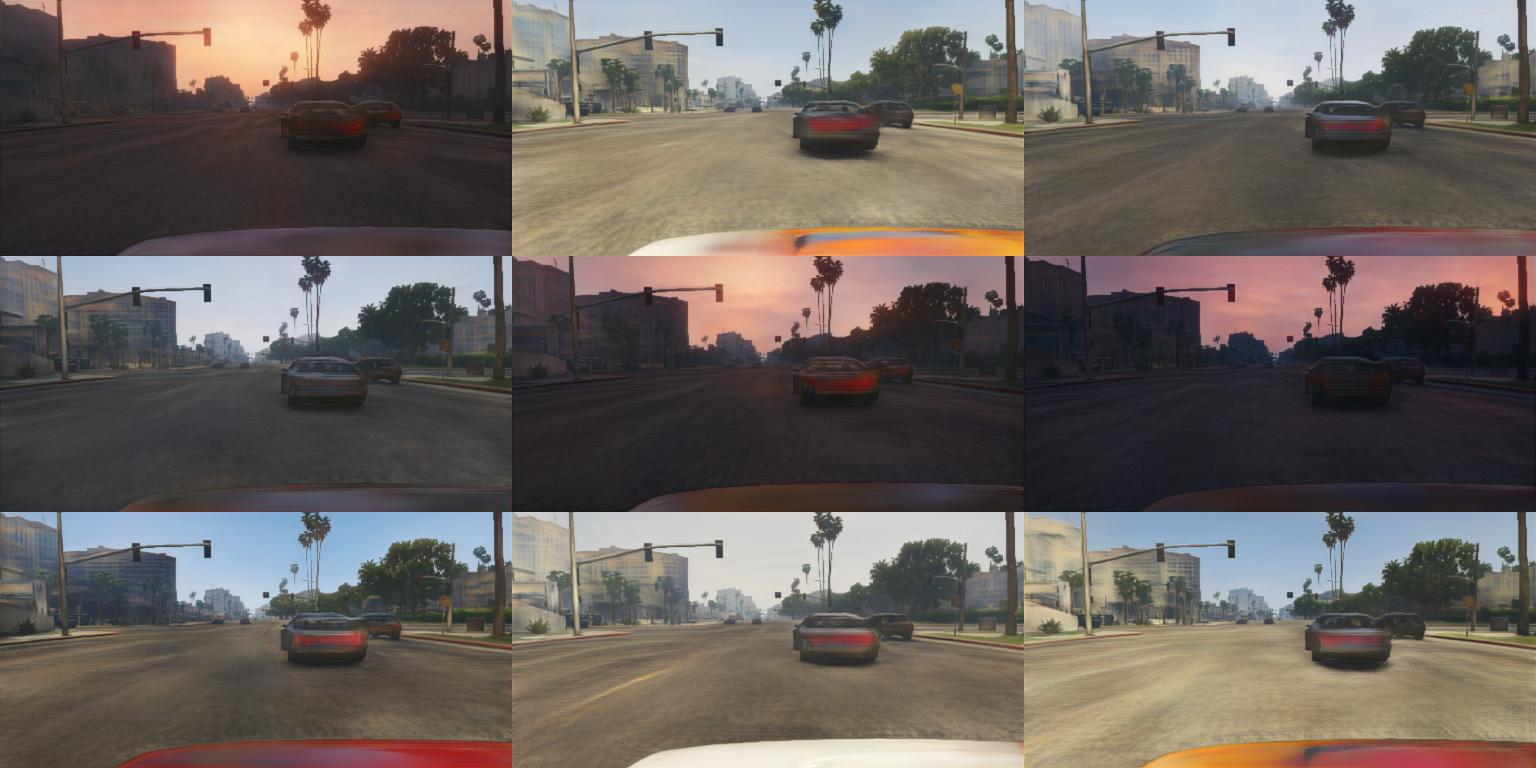}
    \caption{Our model }
    \end{subfigure}
    \caption{Comparison of generated images for the same semantic layout. The bottom-left image in (a) is the input semantic layout and we generate 9 samples for each model. See our website for more samples. }
    \label{fig:samp}
\end{figure*}
\begin{figure*}
    \centering
    \begin{subfigure}{0.47\textwidth}
    \includegraphics[width=\linewidth]{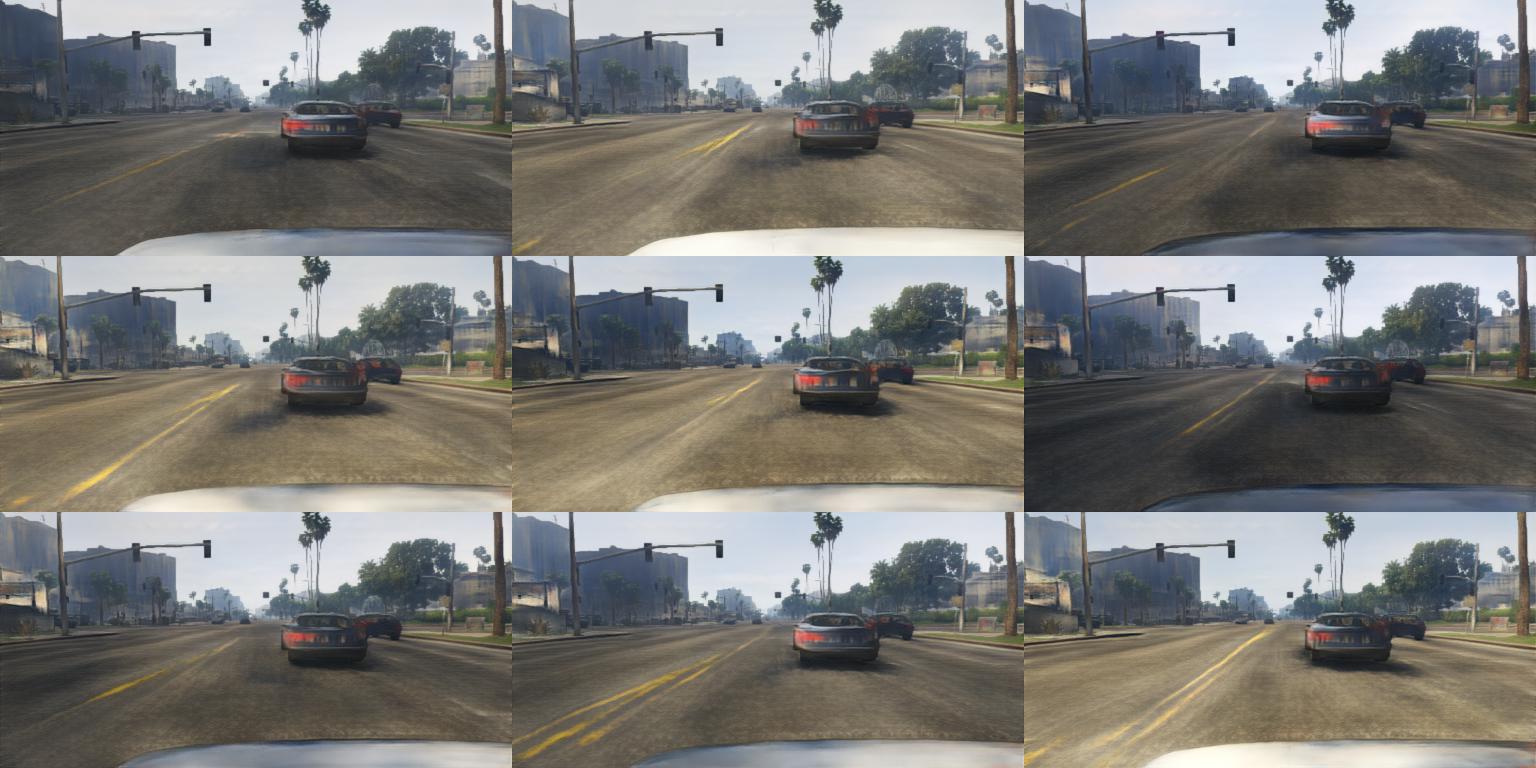}
    \caption{Our model w/o the noise encoder and rebalancing scheme}
    \end{subfigure}
    \begin{subfigure}{0.47\textwidth}
    \includegraphics[width=\linewidth]{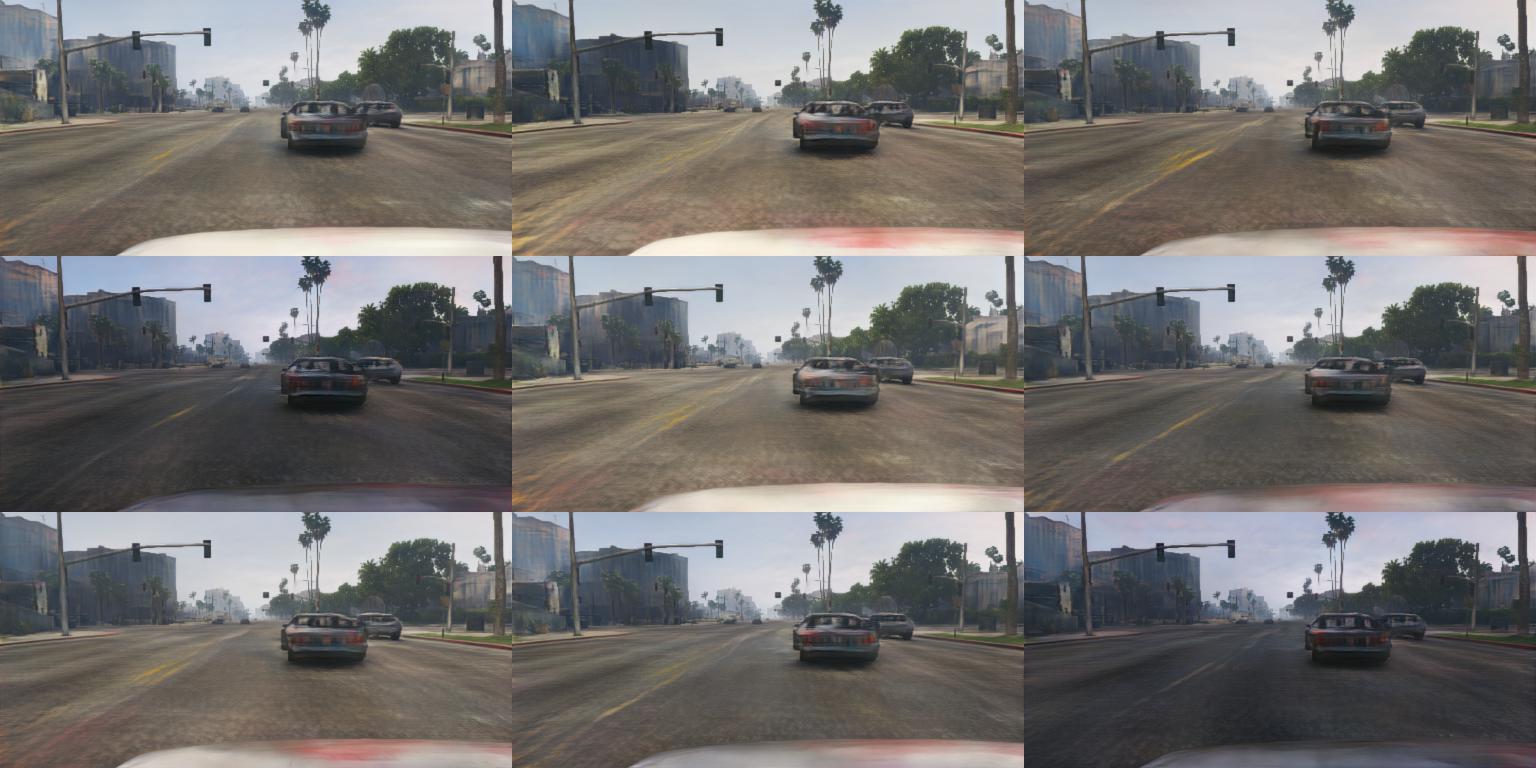}
    \caption{Our model w/o the noise encoder}
    \end{subfigure}
    \begin{subfigure}{0.47\textwidth}
    \includegraphics[width=\linewidth]{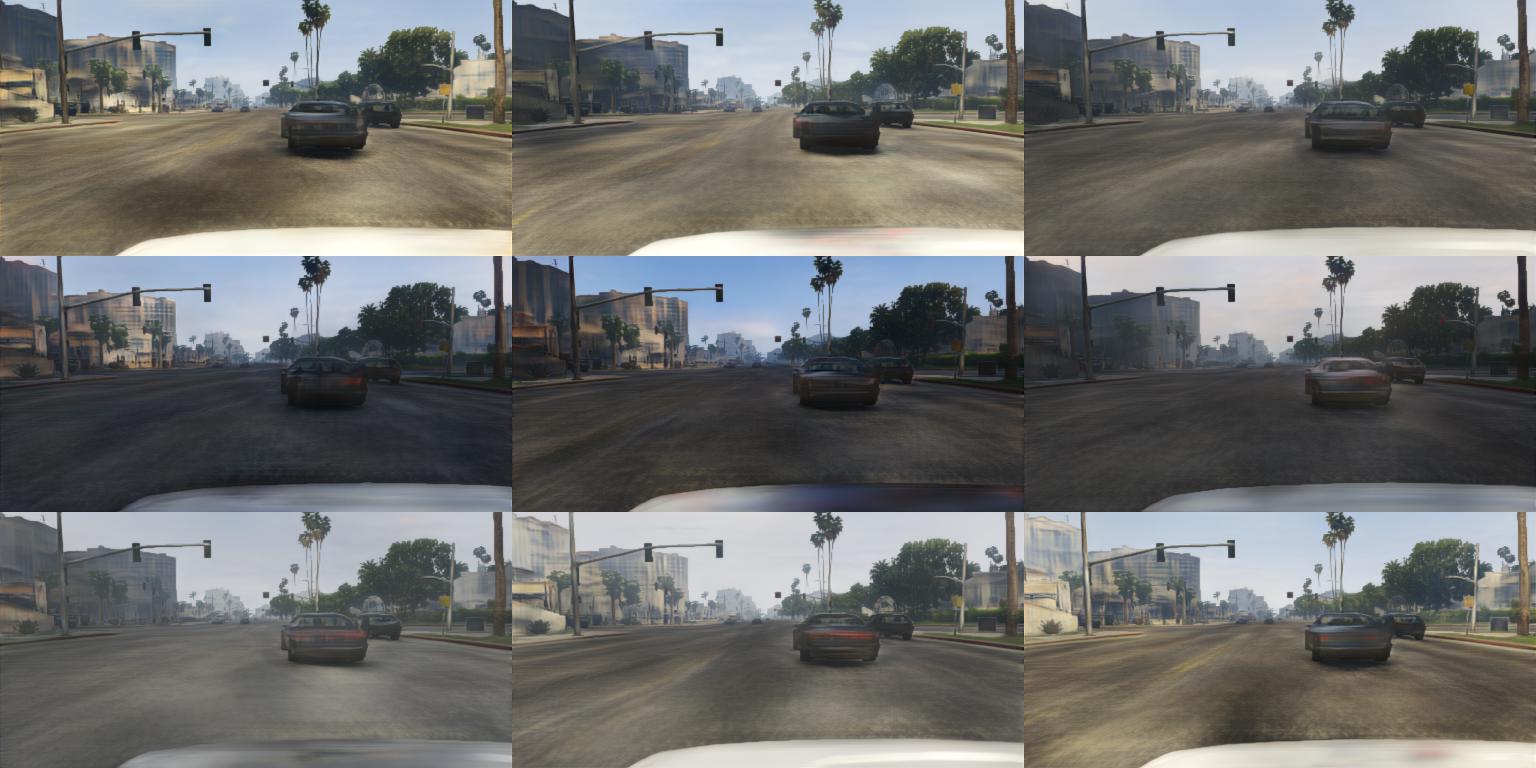}
    \caption{Our model w/o the rebalancing scheme}
    \end{subfigure}
    \begin{subfigure}{0.47\textwidth}
    \includegraphics[width=\linewidth]{fig/aours.jpg}
    \caption{Our model }
    \end{subfigure}
    \caption{Ablation study using the same semantic layout as Fig. \ref{fig:samp}.}
    \label{fig:ablation}
\end{figure*}

\begin{figure*}
    \centering
    \includegraphics[width=0.87\linewidth]{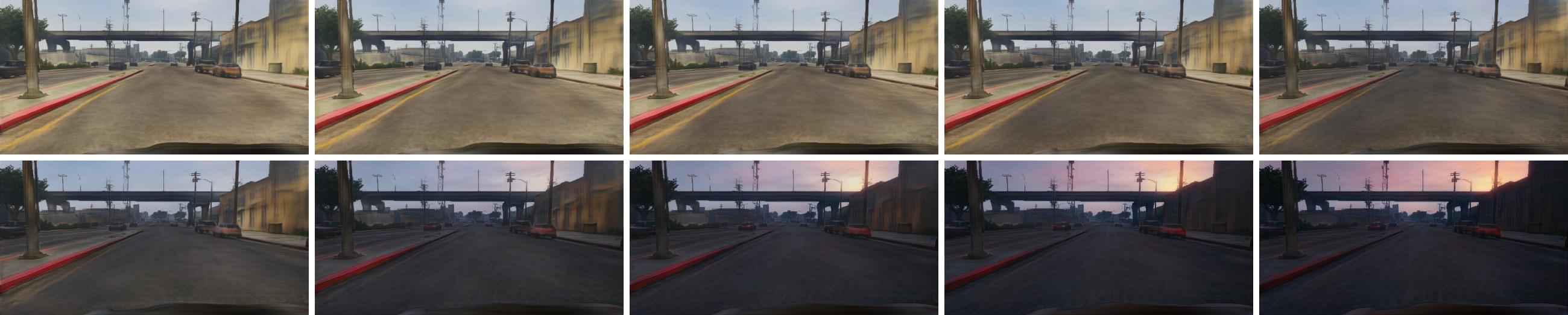}
    \caption{Images generated by interpolating between latent noise vectors. See our website for videos showing the effect of interpolations. }
    \label{fig:interp}
\end{figure*}

\begin{figure*}[htbp]
    \centering
    \begin{subfigure}{0.17\textwidth}
    \includegraphics[width=\linewidth]{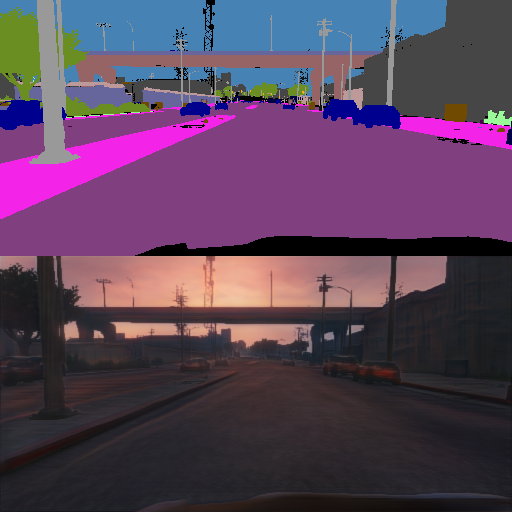}
    \caption{}
    \end{subfigure}
    \begin{subfigure}{0.17\textwidth}
    \includegraphics[width=\linewidth]{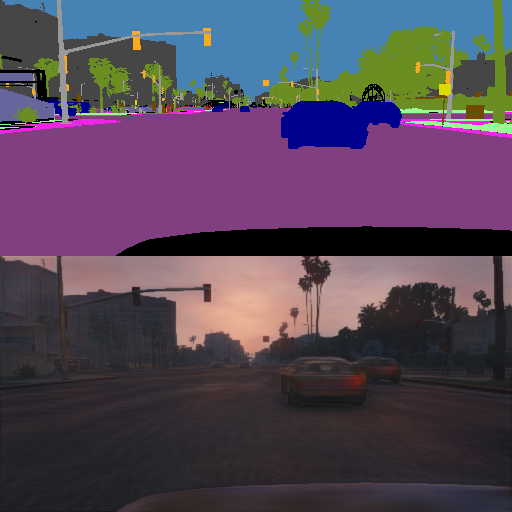}
    \caption{}
    \end{subfigure}
    \begin{subfigure}{0.17\textwidth}
    \includegraphics[width=\linewidth]{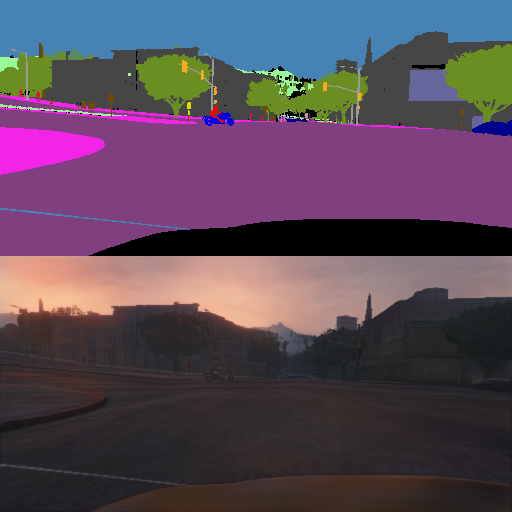}
    \caption{}
    \end{subfigure}
    \begin{subfigure}{0.17\textwidth}
    \includegraphics[width=\linewidth]{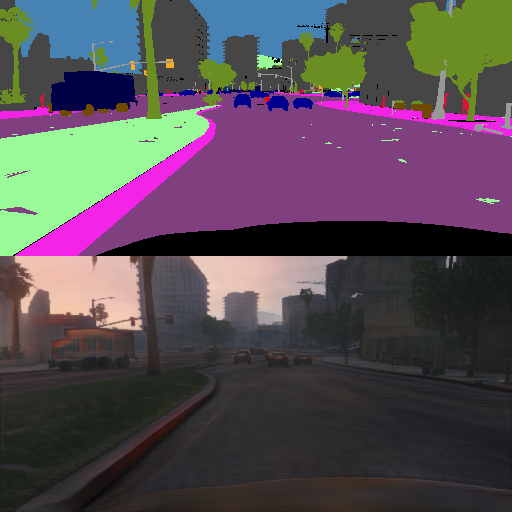}
    \caption{}
    \end{subfigure}
    \begin{subfigure}{0.17\textwidth}
    \includegraphics[width=\linewidth]{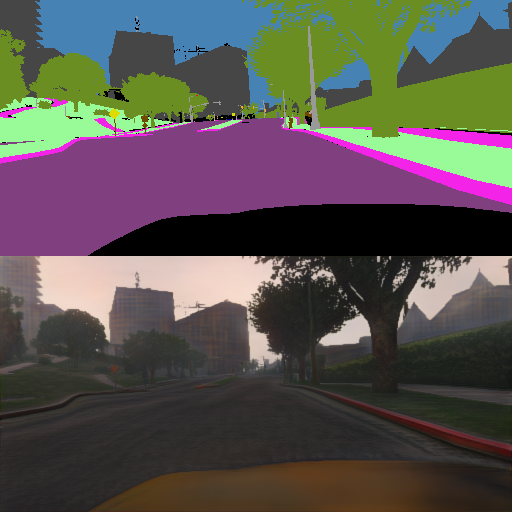}
    \caption{}
    \end{subfigure}
    \caption{Style consistency with the same random vector. (a) is the original input-output pair. We use the same random vector used in (a) and apply it to (b),(c),(d) and (e). See our website for more examples. }
    \label{fig:cons}
\end{figure*}
\subsection{Dataset}
The choice of dataset is very important for multimodal conditional image synthesis. The most common dataset in the unimodal setting is the Cityscapes dataset \cite{Cordts2016Cityscapes}. However, it is not suitable for the multimodal setting because most images in the dataset are taken under similar weather conditions and time of day and the amount of variation in object colours is limited. This lack of diversity limits what any multimodal method can do. On the other hand, the GTA-5 dataset~\cite{Richter_2016_ECCV}, has much greater variation in terms of weather conditions and object appearance. To demonstrate this, we compare the colour distribution of both datasets and present the distributiion of hues of both datasets in Figure~\ref{fig:hist}. As shown, Cityscapes is concentrated around a single mode in terms of hue, whereas GTA-5 has much greater variation in hue. Additionally, the GTA-5 dataset includes more 20000 images and so is much larger than Cityscapes. 

\begin{figure}[htbp]
    \centering
    \begin{subfigure}{0.9\linewidth}
    \includegraphics[width=\linewidth]{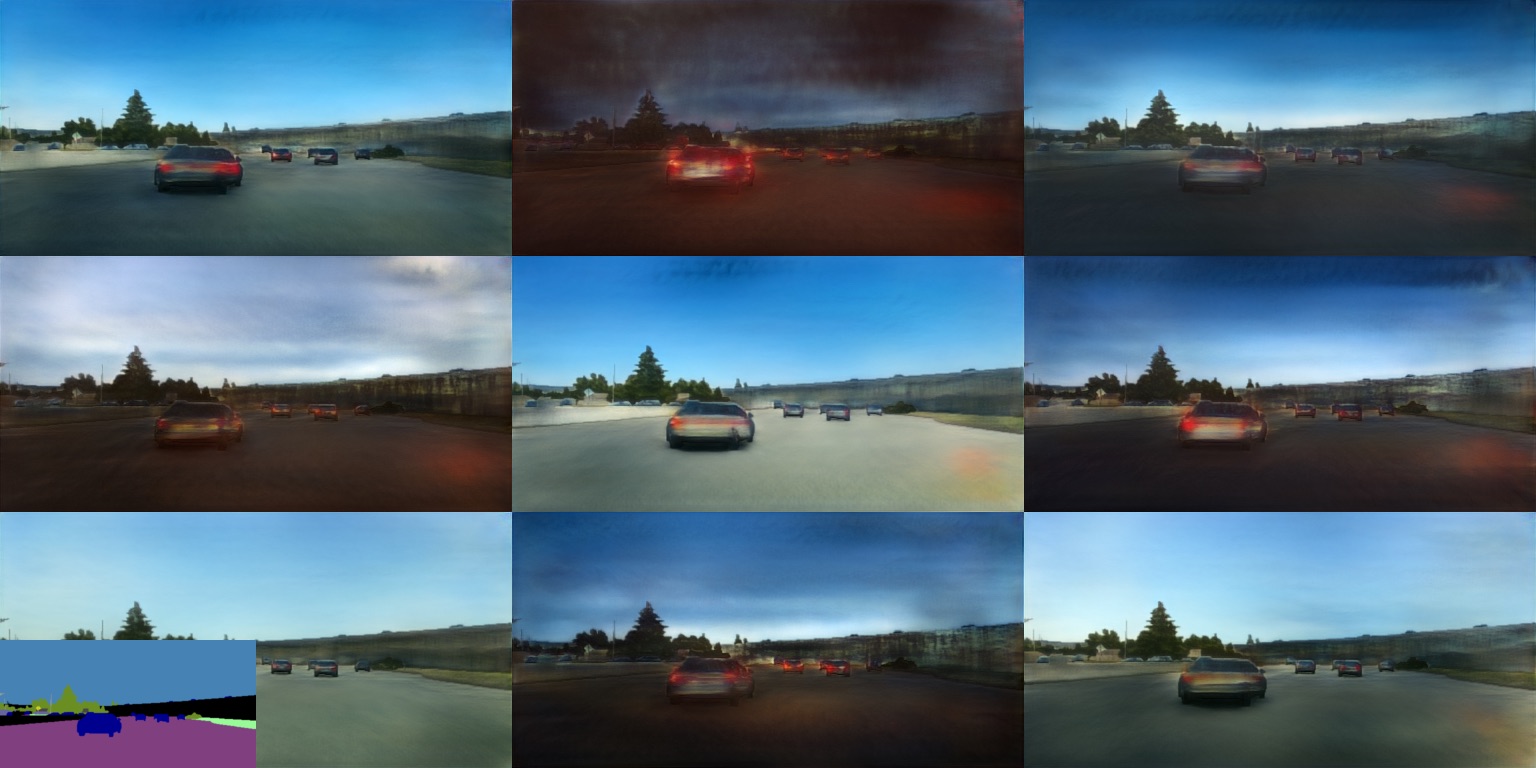}
    \caption{}
    \end{subfigure}
    \begin{subfigure}{0.9\linewidth}
    \includegraphics[width=\linewidth]{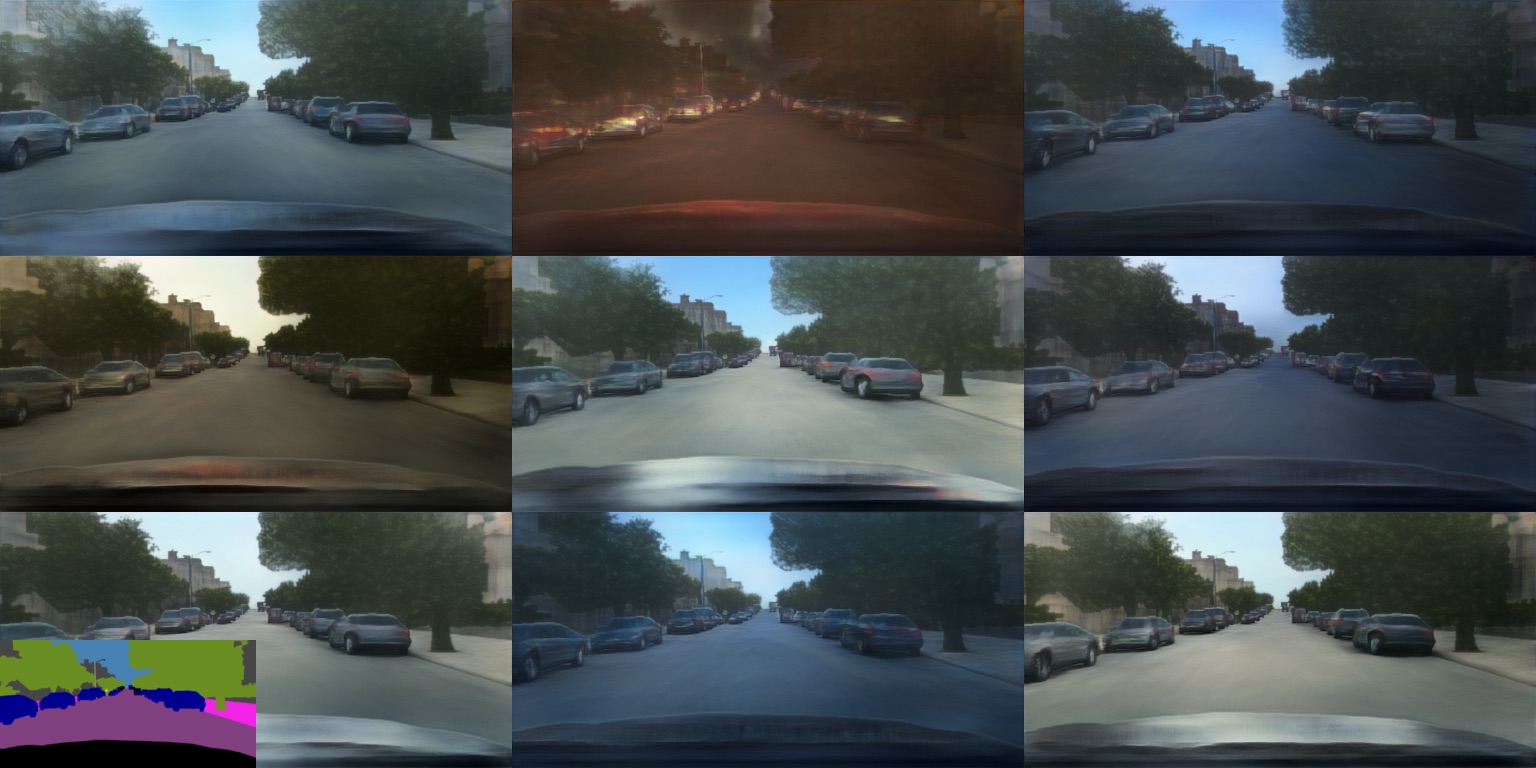}
    \caption{}
    \end{subfigure}
    \caption{Images generated using our method on the BDD100K~\cite{yu2018bdd100k} dataset.}
    \label{fig:bdd}
\end{figure}

Furthermore, to show the generalizability of our approach and its applicability to real-world datasets, we train on the BDD100K~\cite{yu2018bdd100k} dataset and show results in Fig. \ref{fig:bdd}.

\subsection{Experimental Setting}
We train our model on 12403 training images and evaluate on the validation set (6383 images). Due to computational resource limitations, we conduct experiments at the $256 \times 512$ resolution. We add 10 noise channels and set the hyperparameters shown in Algorithm \ref{alg:A} to the following values: $|S|=400$, $m=10$, $K=10000$, $|\Hat{S}|=1$ and $\eta=1e-5$. 

The leading method for image synthesis from semantic layouts in the multimodal setting is the CRN \cite{chen2017photographic} with diversity loss that generates nine different images for each semantic segmentation map and is the baseline that we compare to.

\subsection{Quantitative Comparison}
Quantitative comparison aims to quantitatively compare the diversity as well as quality of the images generated by our model and CRN.

\paragraph{Diversity Evaluation} We measure the diversity of each method by generating 40 pairs of output images for each of 100 input semantic layouts from the test set. We then compute the average distance between each pair of output images for each given input semantic layout, which is then averaged over all input semantic layouts. The distance metric we use is LPIPS~\cite{zhang2018unreasonable}, which is designed to measure perceptual dissimilarity. The results are shown in Table \ref{tab:lpips}. As shown, the proposed method outperforms the baselines by a large margin. We also perform an ablation study and find that the proposed method performs better than variants that remove the noise encoder or the rebalancing scheme, which demonstrates the value of each component of our method. 

\paragraph{Image Quality Evaluation}
We now evaluate the generated image quality by human evaluation. Since it is difficult for humans to compare images with different styles, we selected the images that are closest to the ground truth image in $\ell_1$ distance among the images generated by CRN and our method. We then asked 62 human subjects to evaluate the images generated for 20 semantic layouts. For each semantic layout, they were asked to compare the image generated by CRN to the image generated by our method and judge which image exhibited more obvious synthetic patterns. The result is shown in Table \ref{tab:human}.

\begin{figure}
    \centering
    \begin{subfigure}{0.23\textwidth}
    \includegraphics[width=\linewidth]{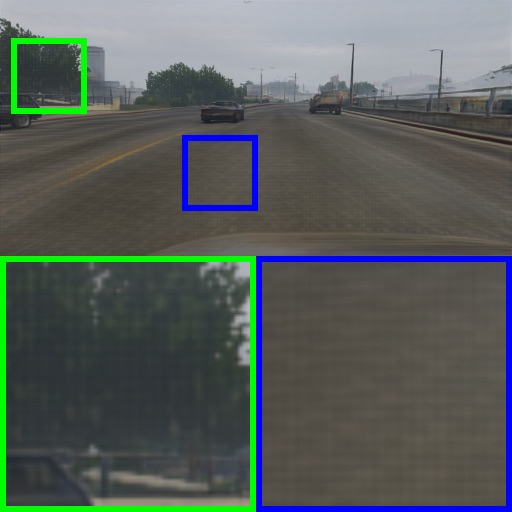}
    \caption{CRN}
    \end{subfigure}
    \begin{subfigure}{0.23\textwidth}
    \includegraphics[width=\linewidth]{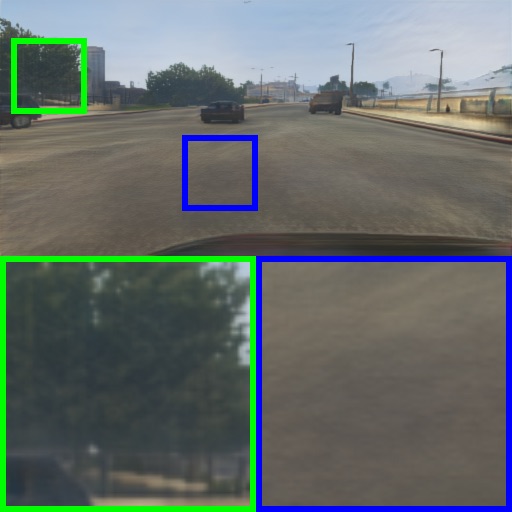}
    \caption{Our model}
    \end{subfigure}
    \caption{Comparison of artifacts in generated images.}
    \label{fig:arti}
\end{figure}
\begin{table}
    \centering
    \footnotesize
\begin{tabular}{cc}
\toprule
     Model& LPIPS score\\
     \midrule
     CRN&$0.11$\\
     CRN+noise&$0.12$\\
     Ours w/o noise encoder&$0.10$\\
     Ours w/o rebalancing scheme&$0.17$\\
     Ours&$\textbf{0.19}$\\
     \bottomrule
\end{tabular}
\caption{LPIPS score. We show the average perceptual distance of different models (including ablation study) and our proposed model gained the highest diversity.}
\label{tab:lpips}
\end{table}
\begin{table}
    \centering
    \footnotesize
\begin{tabular}{cc}
\toprule
     & \% of Images Containing More Artifacts \\
     \midrule
     CRN&$0.636\pm 0.242$\\
     Our method&$\textbf{0.364}\pm 0.242$\\
     \bottomrule
\end{tabular}
\caption{Average percentage of images that are judged by humans to exhibit more obvious synthetic patterns. Lower is better. }
\label{tab:human}
\end{table}
\subsection{Qualitative Evaluation}
A qualitative comparison is shown in Fig. \ref{fig:samp}. We compare to three baselines, BicycleGAN~\cite{zhu2017toward}, Pix2pix-HD with input noise~\cite{wang2017high} and CRN. As shown, Pix2pix-HD generates almost identical images, BicycleGAN generates images with heavy distortions and CRN generates images with little diversity. In comparison, the images generated by our method are diverse and do not suffer from distortions. We also perform an ablation study in Fig. \ref{fig:ablation}, which shows that each component of our method is important. In the supplementary material, we include results of the baselines with the proposed rebalancing scheme and demonstrate that, unlike our method, they cannot take advantage of it. 

In addition, our method also generates fewer artifacts compared to CRN, which is especially interesting because the architecture and the distance metric are the same as CRN. As shown in Fig. \ref{fig:arti}, the images generated by CRN has grid-like artifacts which are not present in the images generated by our method. More examples generated by our model are shown in the supplementary material.

\paragraph{Interpolation}
We also perform linear interpolation of latent vectors to evaluate the semantic structure of the learned latent space. As shown in \ref{fig:interp}, by interpolating between the noise vectors corresponding to generated images during daytime and nighttime respectively, we obtain a smooth transition from daytime to nighttime. 
This suggests that the learned latent space is sensibly ordered and captures the full range of variations along the time-of-day axis. More examples are available in the supplementary material.

\begin{figure}[htbp]
    \centering
    \begin{subfigure}{0.2\textwidth}
    \includegraphics[width=\linewidth]{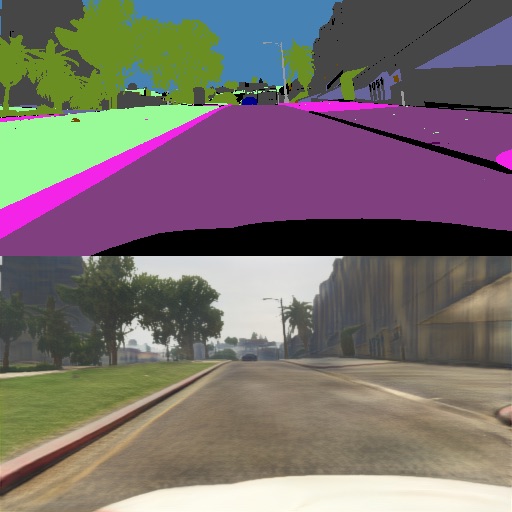}
    \caption{}
    \end{subfigure}
    \begin{subfigure}{0.2\textwidth}
    \includegraphics[width=\linewidth]{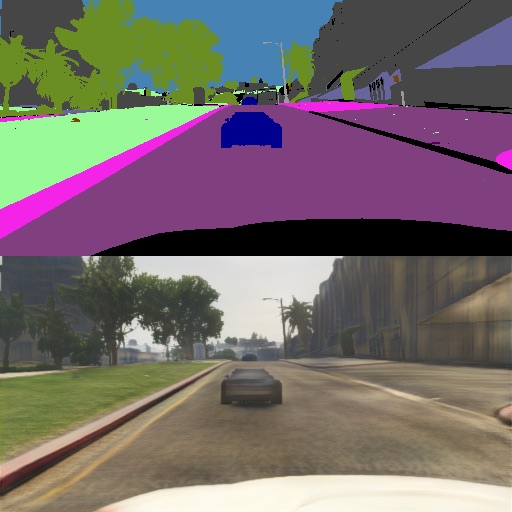}
    \caption{}
    \end{subfigure}
    \begin{subfigure}{0.2\textwidth}
    \includegraphics[width=\linewidth]{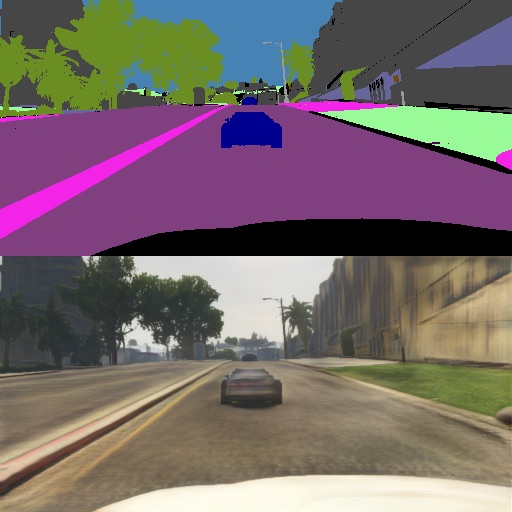}
    \caption{}
    \end{subfigure}
    \begin{subfigure}{0.2\textwidth}
    \includegraphics[width=\linewidth]{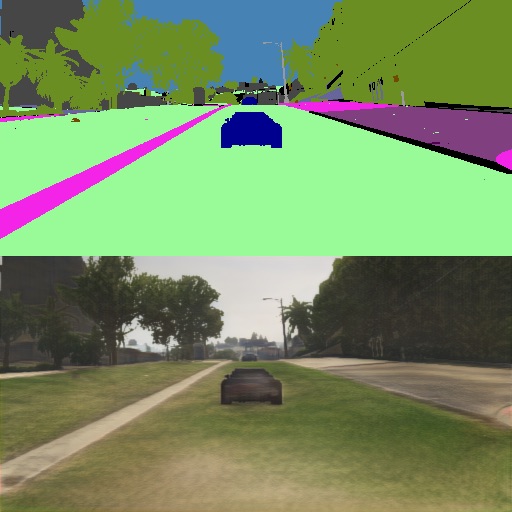}
    \caption{}
    \end{subfigure}
    \caption{Scene editing. (a) is the original input semantic map and the generated output. (b) adds a car on the road. (c) changes the grass on the left to road and change the side walk on the right to grass. (d) deletes our own car, changes the building on the right to tree and changes all road to grass.}
    \label{fig:edit}
\end{figure}

\paragraph{Scene Editing}
A successful method for image synthesis from semantic layouts enables users to manually edit the semantic map to synthesize desired imagery. One can do this simply by adding/deleting objects or changing the class label of a certain object. In Figure \ref{fig:edit} we show several such changes. Note that all four inputs use the same random vector; as shown, the images are highly consistent in terms of style, which is quite useful because the style should remain the same after editing the layout. We further demonstrate this in Fig. \ref{fig:cons} where we apply the random vector used in (a) to different segmentation maps in (b),(c),(d),(e) and the style is preserved across the different segmentation maps.

\section{Conclusion}
We presented a new method based on IMLE for multimodal image synthesis from semantic layout. Unlike prior approaches, our method can generate arbitrarily many images for the same semantic layout and is easy to train. We demonstrated that our method can generate more diverse images with fewer artifacts compared to the leading approach~\cite{chen2017photographic}, despite using the same architecture.
In addition, our model is able to learn a sensible latent space of noise vectors without supervision. We showed that by taking the interpolations between noise vectors, our model can generate continuous changes. At the same time, using the same noise vector across different semantic layouts result in images of consistent style.

{\small
\bibliographystyle{ieee_fullname}
\bibliography{diverse}
}
\clearpage

\onecolumn
\title{Diverse Image Synthesis from Semantic Layouts via Conditional IMLE \\ \vspace{5pt} \large Supplementary Material}

\author{Ke Li$^{\ast}$\\
UC Berkeley\\
{\tt\small ke.li@eecs.berkeley.edu}
\and
Tianhao Zhang$^{\ast}$\\
Nanjing University\\
{\tt\small bryanzhang@smail.nju.edu.cn}
\and
Jitendra Malik\\
UC Berkeley\\
{\tt\small malik@eecs.berkeley.edu}
}
\maketitle

\begin{appendices}

\section{Implementation Details}

\paragraph{One-Hot Embedding of Semantic Layout}

The segmentation map $\mathbf{x}$ that is fed as input to the neural net $T_\theta$ is encoded in a one-hot fashion. Specifically, each pixel in the segmentation map is represented as a one-hot encoding of the semantic category it belongs to, that is
\begin{equation*}
    \mathbf{x}^{i,j,p} = \begin{cases}
    1, \text{pixel at location } (i,j) \text{ belongs to class } p\\
    0, \text{otherwise}
    \end{cases}
\end{equation*}

\paragraph{Distance Metric} For the distance metric $\mathcal{L}(\cdot,\cdot)$, we use the loss function used by CRN, namely a perceptual loss function based on VGG-19 features~\cite{simonyan2014very}:
\begin{equation}\label{eq:loss}
    \mathcal{L}(\mathbf{y}, \widetilde{\mathbf{y}}) = \sum_{i=1}^l\lambda_i\left\Vert\Phi_i(\mathbf{y})-\Phi_i(\widetilde{\mathbf{y}})\right\Vert_1
\end{equation}

Here $\Phi_1(\cdot),\cdots,\Phi_l(\cdot)$ represents the feature outputs of the following layers in VGG-19: ``conv1\_2'', ``conv2\_2'', ``conv3\_2'', ``conv4\_2'', and ``conv5\_2''. Hyperparameters $\{\lambda_i\}_{i=1}^l$ are set such that the loss of each layer makes the same contribution to the total loss. We use this loss function as the distance metric in the IMLE objective.

\paragraph{Dataset Rebalancing} We first rebalance the dataset to increase the chance of rare images being sampled when populating the training batch $S$. To this end, for each training image, we calculate the average colour of each category in that image. Then for each category, we estimate the distribution of the average colour of the category over different training images using a Gaussian kernel density estimate (KDE). 

More concretely, we compute the average colour of each semantic category $p$ for each image $k$, which is a three-dimensional vector:
\begin{equation*}
    \mathbf{c}_{k}(p)= \frac{\sum_{i=1}^{h}\sum_{j=1}^{w} \mathbf{1}\left[\mathbf{x}_k^{i,j}=p\right] \mathbf{y}_k^{i,j}}{\sum_{i=1}^{h}\sum_{j=1}^{w} \mathbf{1}\left[\mathbf{x}_k^{i,j}=p\right]} = \frac{\sum_{i=1}^{h}\sum_{j=1}^{w} \mathbf{x}_k^{i,j,p} \mathbf{y}_k^{i,j}}{\sum_{i=1}^{h}\sum_{j=1}^{w} \mathbf{x}_k^{i,j,p}}
\end{equation*}

For each category $p$, we consider the set of average colours for that category in all training images, i.e.: $\{\mathbf{c}_{k}(p) \vert k \in \{1,\ldots,n\}\text{ such that class }p\text{ appears in }\mathbf{x}_k\}$. We then fit a Gaussian kernel density estimate to this set of vectors and obtain an estimate of the distribution of average colours of category $p$. Let $D_p(\cdot)$ denote the estimated probability density function (PDF) for category $p$. We define the \textit{rarity score} of category $p$ in the $k^{\mathrm{th}}$ training image as follows:
\begin{equation*}
    R_p(k) = \begin{cases}
    \frac{1}{D_p(\mathbf{c}_{k}(p))} &\text{class } p \text{ appears in } \mathbf{x}_k\\
    0 &\text{otherwise}
    \end{cases}
\end{equation*}

When populating training batch $S$, we allocate a portion of the batch to each of the top five categories that have the largest overall area across the dataset. For each category, we sample training images with a probability in proportion of their rarity scores. Effectively, we upweight images containing objects with rare appearance. 

The rationale for selecting the categories with the largest areas is because they tend to appear more frequently and be visually more prominent. If we were to allocate a fixed portion of the batch to rare categories, we would risk overfitting to images containing those categories.

\paragraph{Loss Rebalancing} The same training image can contain both common and rare objects. Therefore, we modify the loss function so that the objects with rare appearance are upweighted. For each training pair $(\mathbf{x}_k,\mathbf{y}_k)$, we define a rarity score mask $\mathcal{M}^k\in \mathbb{R}^{h\times w\times 1}$:
\begin{equation*}
    \mathcal{M}^k_{i,j} = R_p(k)\quad \text{if pixel }(i,j) \text{ belongs to class }p
\end{equation*}
We then normalize $\mathcal{M}^k$ so that every entry lies in $(0,1]$:
\begin{equation*}
    \widehat{\mathcal{M}}^k = \frac{1}{\max_{i,j} \mathcal{M}^k_{i,j}}\mathcal{M}^k
\end{equation*}
The mask is then used to weight different pixels differently the loss function~\eqref{eq:loss}. Let $\widehat{\mathcal{M}}$ be the normalized rarity score mask associated with the training pair $(\mathbf{x},\mathbf{y})$. The new loss $\mathcal{L}$ becomes:
\begin{equation*}
    \mathcal{L}(\mathbf{y}, \widetilde{\mathbf{y}}) = \sum_{i=1}^l\lambda_l\left\Vert\widehat{\mathcal{M}}_i\circ\left[\Phi_i(\mathbf{y})-\Phi_i(\widetilde{\mathbf{y}})\right]\right\Vert_1
\end{equation*}
Here $\widehat{\mathcal{M}}_i$ is the normalized rarity score mask $\widehat{\mathcal{M}}$ downsampled to match the size of $\Phi_i(\cdot)$, and $\circ$ denotes the element-wise product. 

\newpage
\section{Baselines with Proposed Rebalancing Scheme}

\begin{figure}[h]
    \centering
    \begin{subfigure}{0.45\textwidth}
    \includegraphics[width=\linewidth]{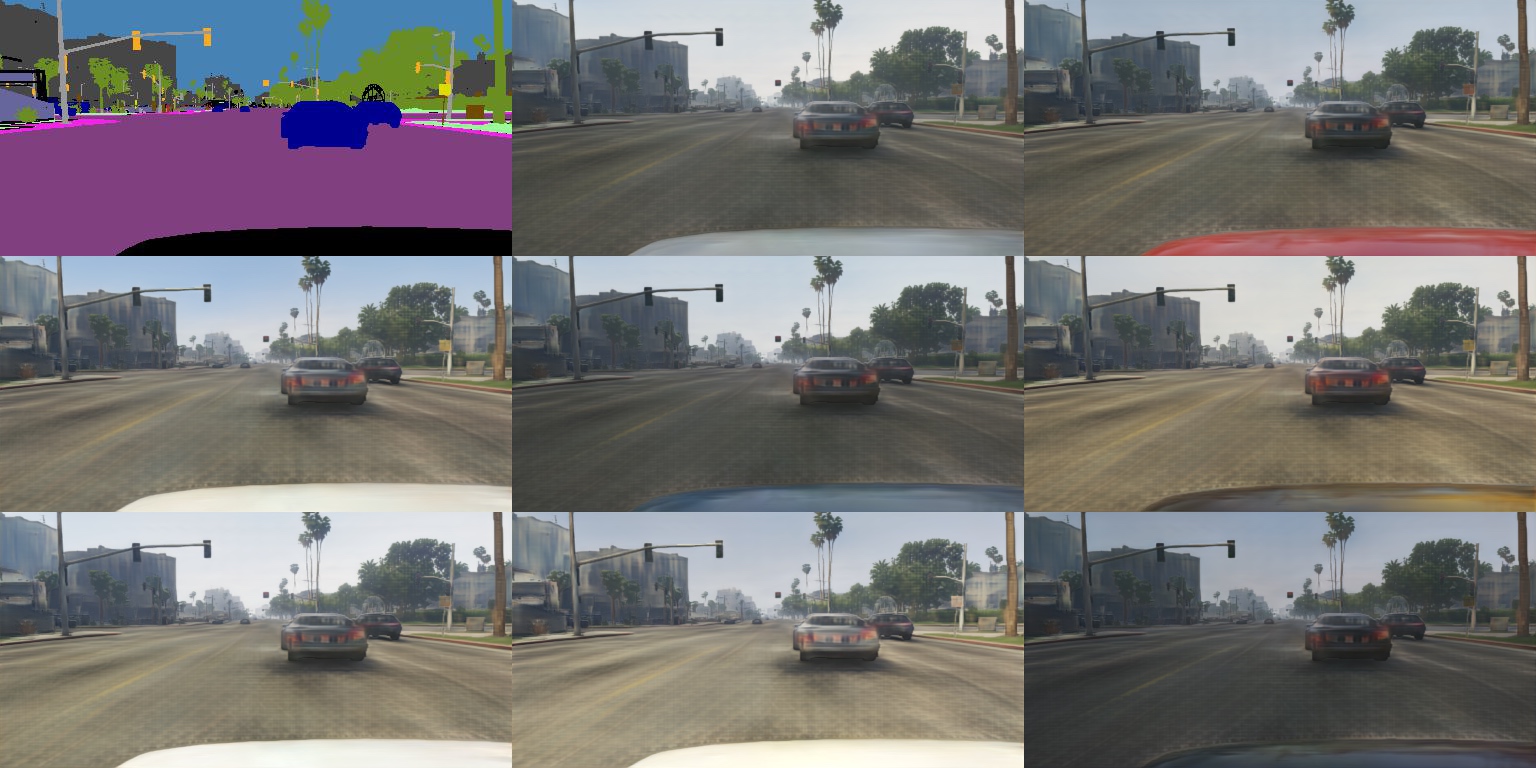}
    \caption{CRN}
    \end{subfigure}
    \begin{subfigure}{0.45\textwidth}
    \includegraphics[width=\linewidth]{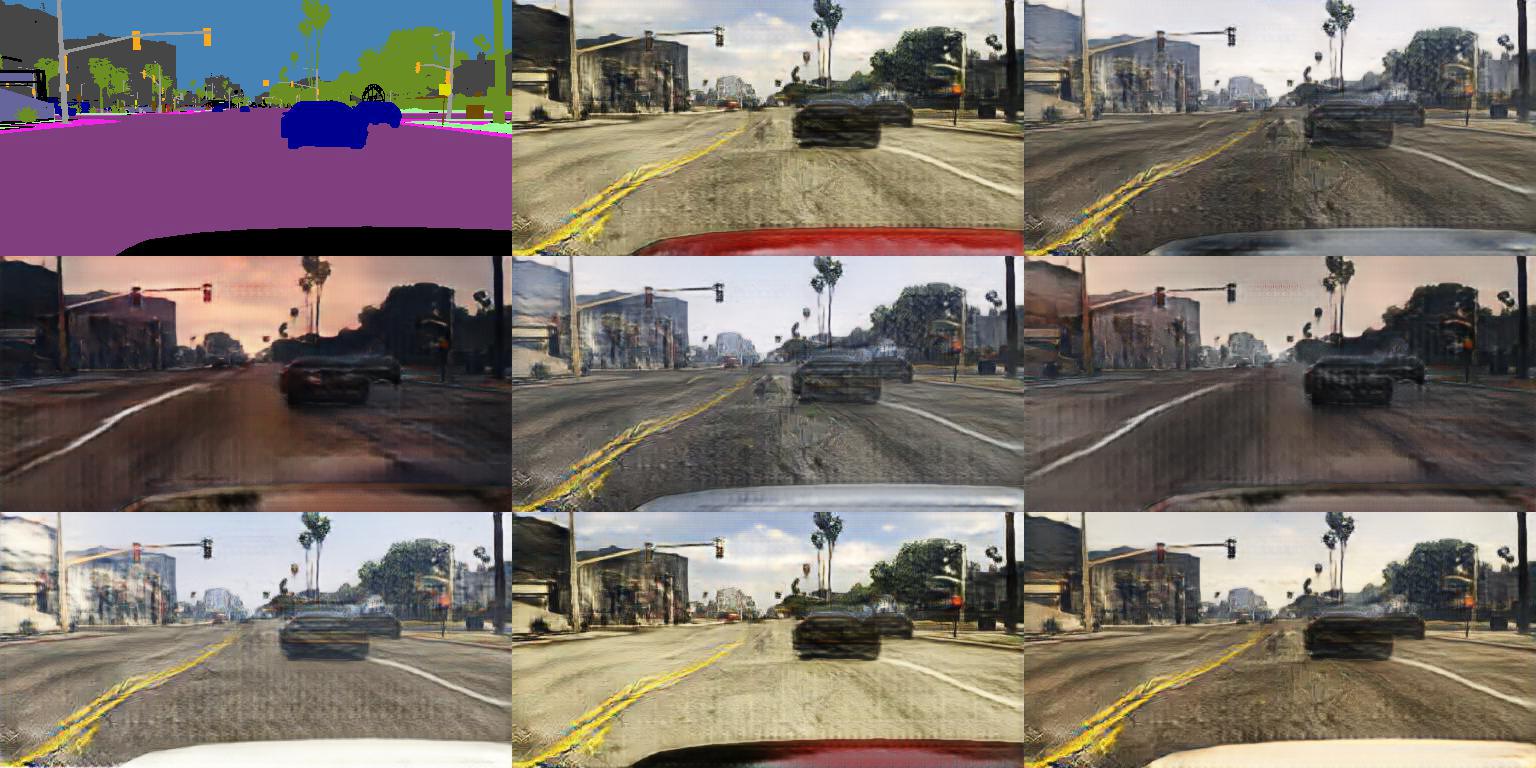}
    \caption{BicycleGAN}
    \end{subfigure}
    \begin{subfigure}{0.45\textwidth}
    \includegraphics[width=\linewidth]{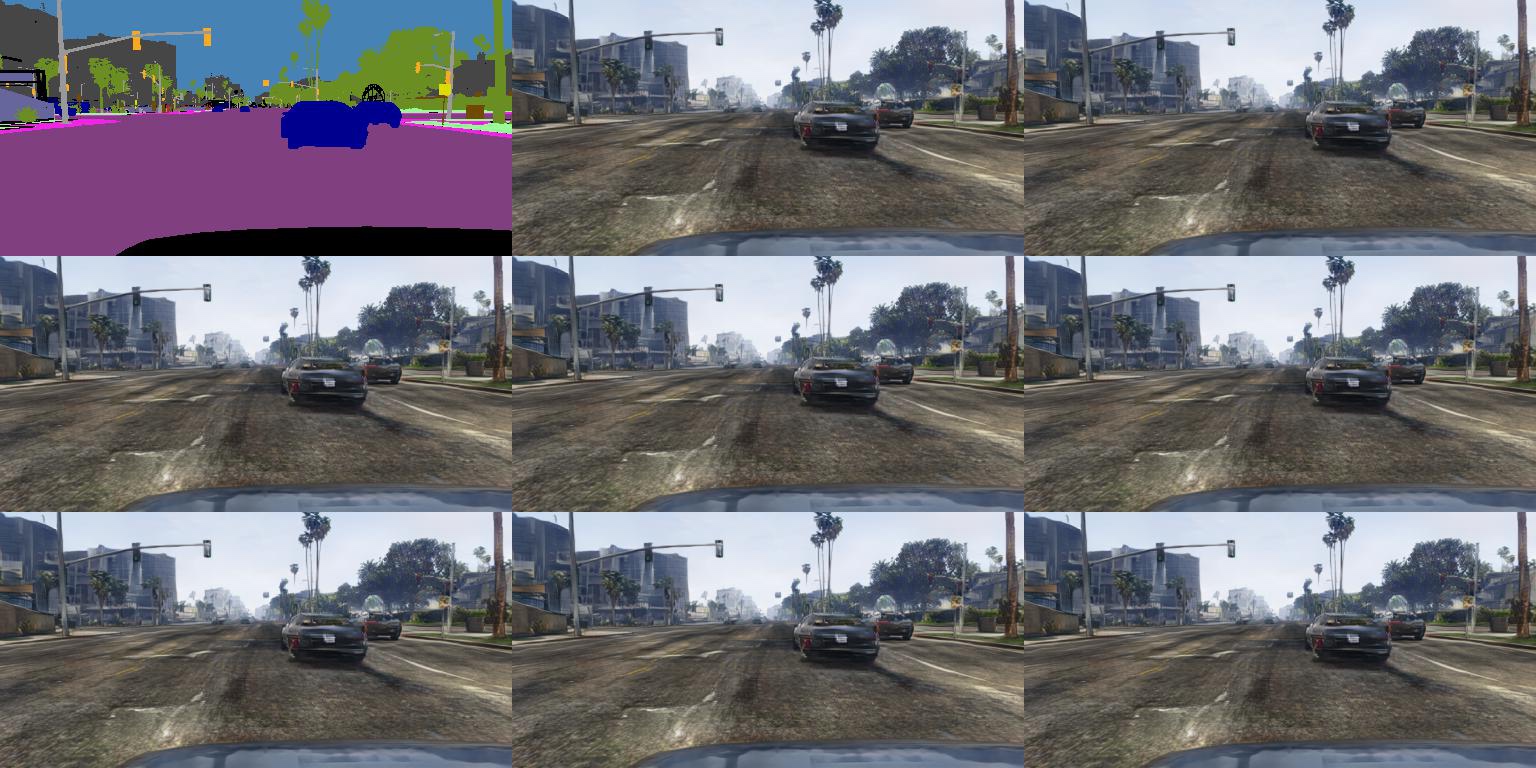}
    \caption{Pix2pix-HD}
    \end{subfigure}
    \begin{subfigure}{0.45\textwidth}
    \includegraphics[width=\linewidth]{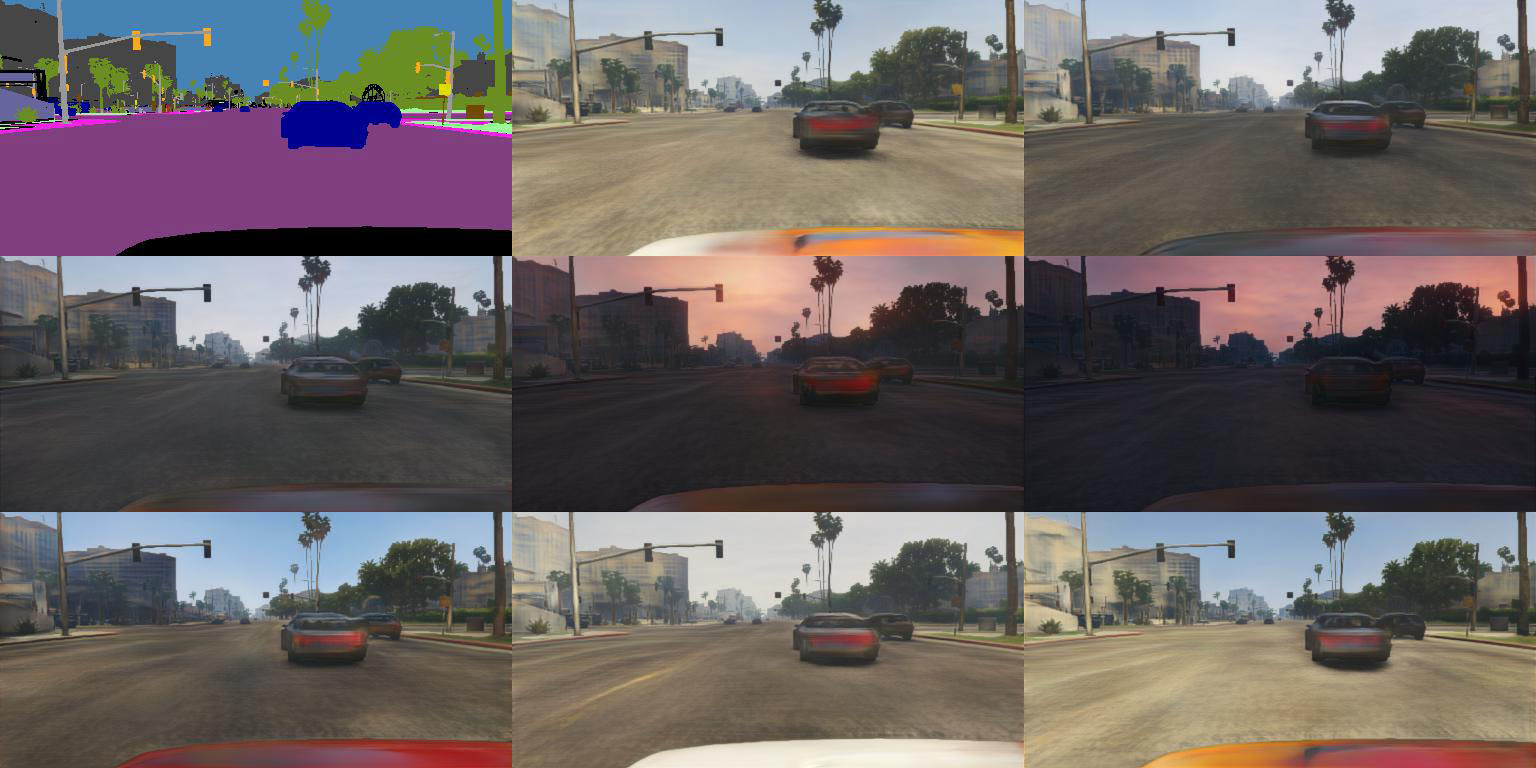}
    \caption{Ours}
    \end{subfigure}
    \caption{Samples generated by baselines with proposed rebalancing scheme, compared to the samples generated by our method. As shown, even with the proposed rebalancing scheme, the samples generated by CRN and Pix2pix-HD exhibit far less diversity than the samples generated by our method, and the samples generated by BicycleGAN are both less diverse and contain more artifacts than the samples generated by our method.}
    \label{fig:rebalance}
\end{figure}

A natural question is whether applying the proposed rebalancing scheme to the baselines would result in a significant improvement in the diversity of generated images. We tried this and found that the diversity is still lacking; the results are shown in Figure~\ref{fig:rebalance}. The LPIPS score of CRN only improves slightly from 0.12 to 0.13 after dataset and loss rebalancing are applied. It still underperforms our method, which achieves a LPIPS score of 0.19. The LPIPS score of Pix2pix-HD showed no improvement after applying dataset rebalancing; it still ignores the latent input noise vector. This suggests that the baselines are not able to take advantage of the rebalancing scheme. On the other hand, our method is able to take advantage of it, demonstrating its superior capability compared to the baselines. 

\newpage

\section{Additional Results}

All videos that we refer to below are available at \url{http://people.eecs.berkeley.edu/~ke.li/projects/imle/scene_layouts}.

\subsection{Video of Interpolations}

\begin{figure*}[h]
    \centering
    \begin{subfigure}{\textwidth}
    \includegraphics[width=\linewidth]{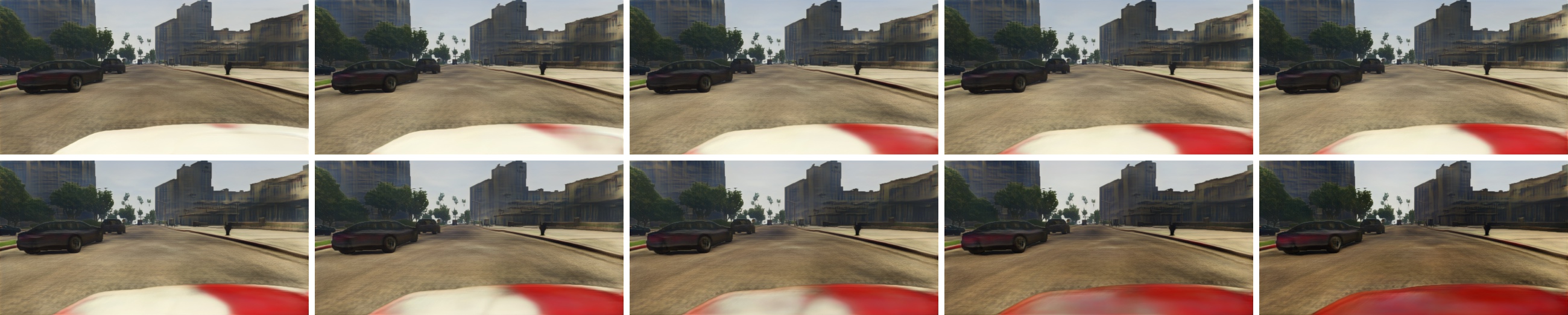}
    \caption{}
    \end{subfigure}
    \begin{subfigure}{\textwidth}
    \includegraphics[width=\linewidth]{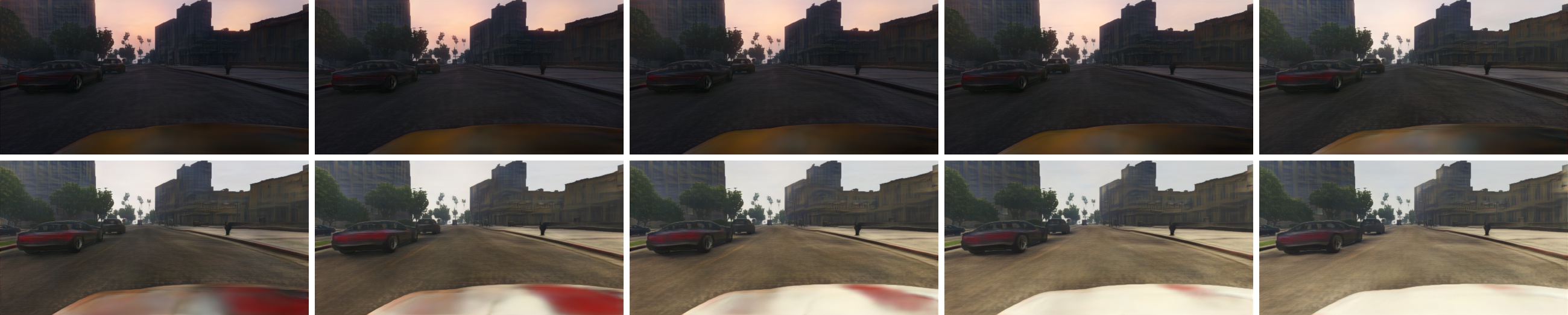}
    \caption{}
    \end{subfigure}
    \caption{\label{fig:interpvid}Frames of video generated by smoothly interpolating between latent noise vectors. }
    
\end{figure*}

We generated a video that shows smooth transitions between different renderings of the same scene. Frames of the generated video are shown in Figure~\ref{fig:interpvid}. 

\newpage
\subsection{Video Generation from Evolving Scene Layouts}

\begin{figure*}[h]
    \centering
    \begin{subfigure}{\textwidth}
    \includegraphics[width=\linewidth]{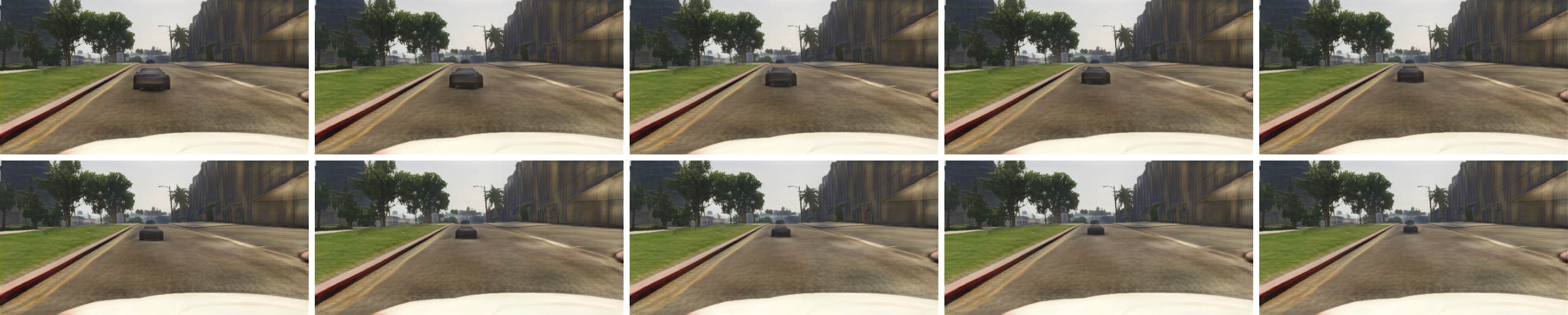}
    \caption{}
    \end{subfigure}
    \begin{subfigure}{\textwidth}
    \includegraphics[width=\linewidth]{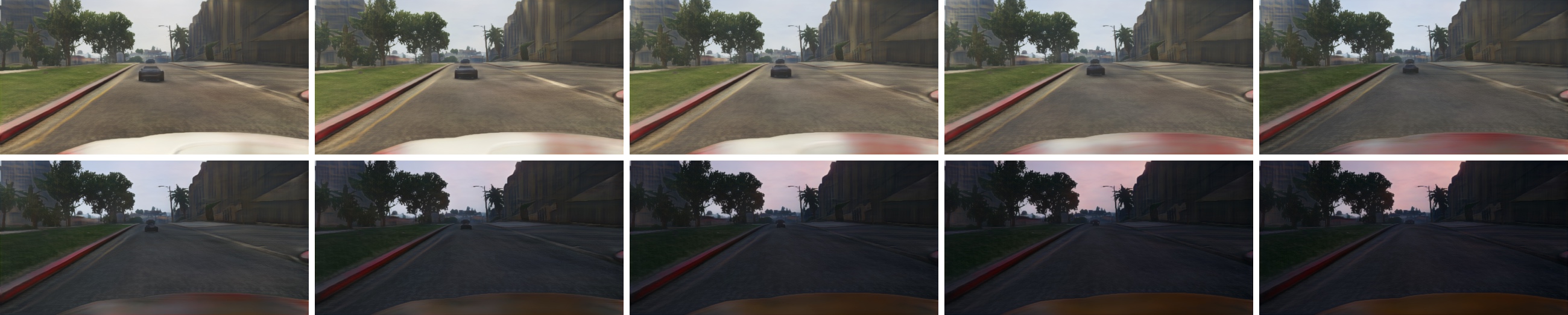}
    \caption{}
    \end{subfigure}
    \caption{\label{fig:vidgen}Frames from two videos of a moving car generated using our method. In both videos, we feed in scene layouts with cars of varying sizes to our model to generate different frames. In (a), we use the same latent noise vector across all frames. In (b), we interpolate between two latent noise vectors, one of which corresponds to a daytime scene and the other to a night time scene. The consistency of style across frames demonstrates that the learned space of latent noise vectors is semantically meaningful and that scene layout and style are successfully disentangled by our model. }
    
\end{figure*}

We generated videos of a car moving farther away from the camera and then back towards the camera by generating individual frames independently using our model with different semantic segmentation maps as input. For the video to have consistent appearance, we must be able to consistently select the same mode across all frames. In Figure~\ref{fig:vidgen}, we show that our model has this capability: we are able to select a mode consistently by using the same latent noise vector across all frames. 

\newpage

\begin{figure*}[h]
    \centering
    \begin{subfigure}{\textwidth}
    \includegraphics[width=\linewidth]{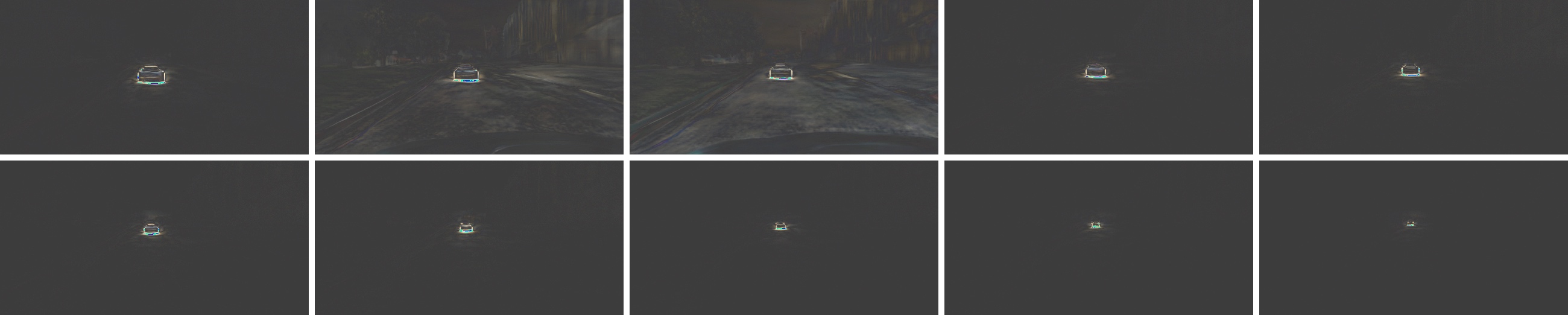}
    \caption{}
    \end{subfigure}
    \begin{subfigure}{\textwidth}
    \includegraphics[width=\linewidth]{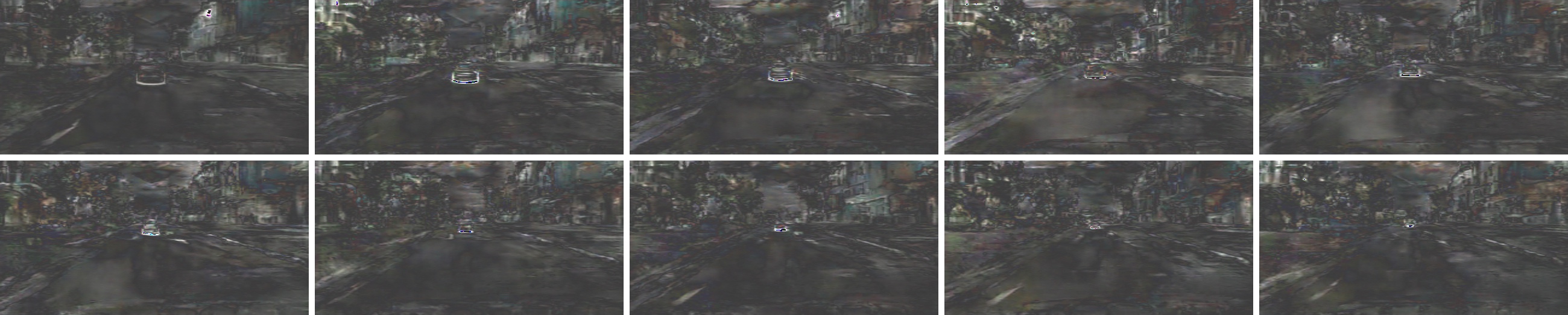}
    \caption{}
    \end{subfigure}
    \caption{\label{fig:vidgendiff}Comparison of the difference between adjacent frames of synthesized moving car video. Darker pixels indicate smaller difference and lighter pixels indicate larger difference. (a) shows results for the video generated by our model. (b) shows results for the video generated by pix2pix~\cite{isola2017image}. }
    
\end{figure*}

Here we demonstrate one potential benefit of modelling multiple modes instead of a single mode. We tried generating a video from the same sequence of scene layouts using pix2pix~\cite{isola2017image}, which only models a single mode. (For pix2pix, we used a pretrained model trained on Cityscapes, which is easier for the purposes of generating consistent frames because Cityscapes is less diverse than GTA-5.) In Figure~\ref{fig:vidgendiff}, we show the difference between adjacent frames in the videos generated by our model and pix2pix. As shown, our model is able to generate consistent appearance across frames (as evidenced by the small difference between adjacent frames). On the other hand, pix2pix is not able to generate consistent appearance across frames, because it arbitrarily picks a mode to generate and does not permit control over which mode it generates. 

\newpage
\section{More Generated Samples}

\begin{figure*}[ht]
    \centering
    \begin{subfigure}{\textwidth}
    \includegraphics[width=\linewidth]{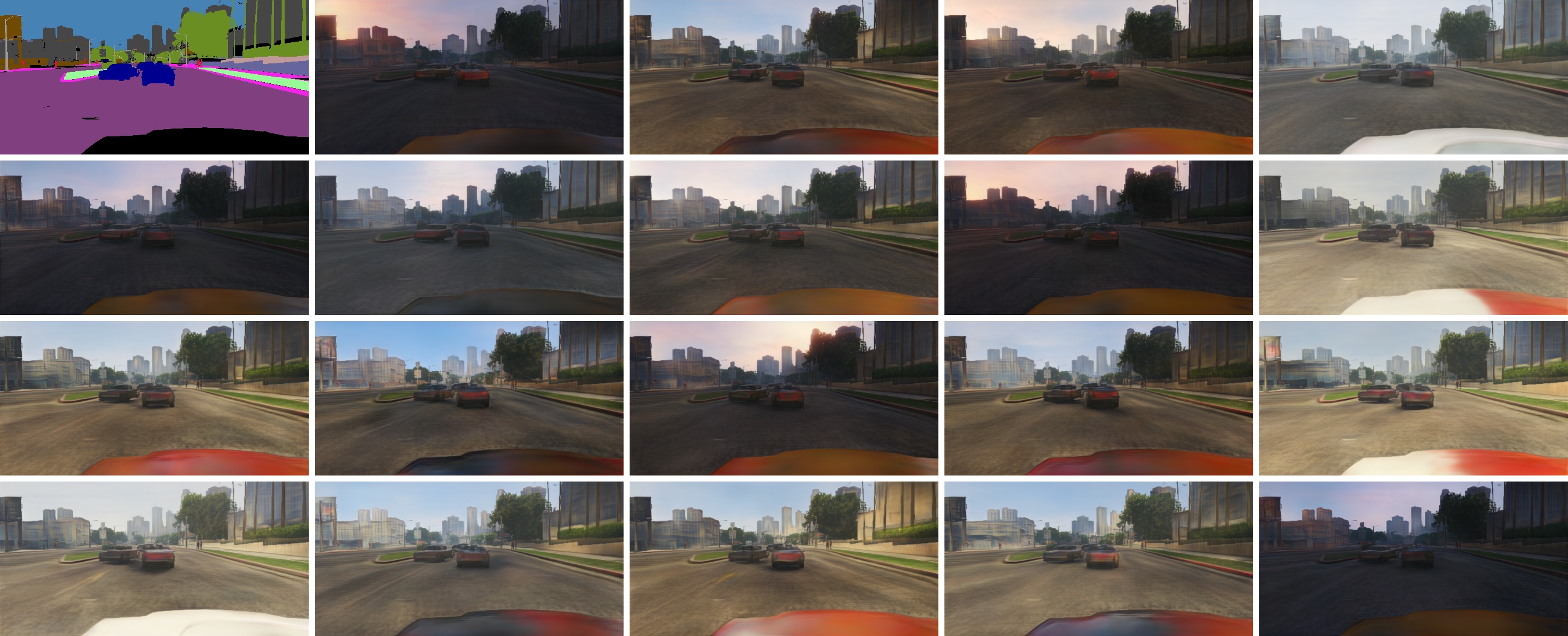}
    \caption{}
    \end{subfigure}
    \begin{subfigure}{\textwidth}
    \includegraphics[width=\linewidth]{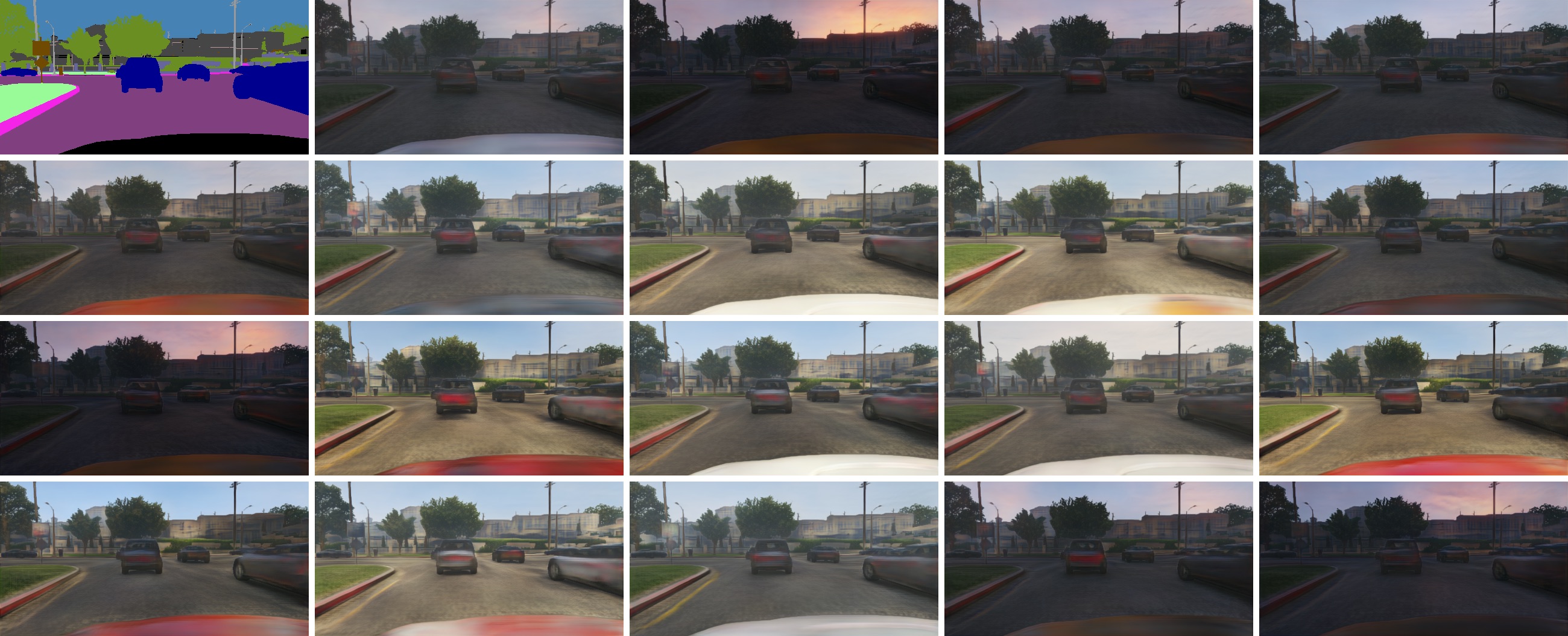}
    \caption{}
    \end{subfigure}
    \caption{Samples generated by our model. The image at the top-left corner is the input semantic layout and the other 19 images are samples generated by our model conditioned on the same semantic layout.}
    \label{fig:eg20_more}
\end{figure*}
\newpage
\begin{figure*}[ht]
    \centering
    \begin{subfigure}{\textwidth}
    \includegraphics[width=\linewidth]{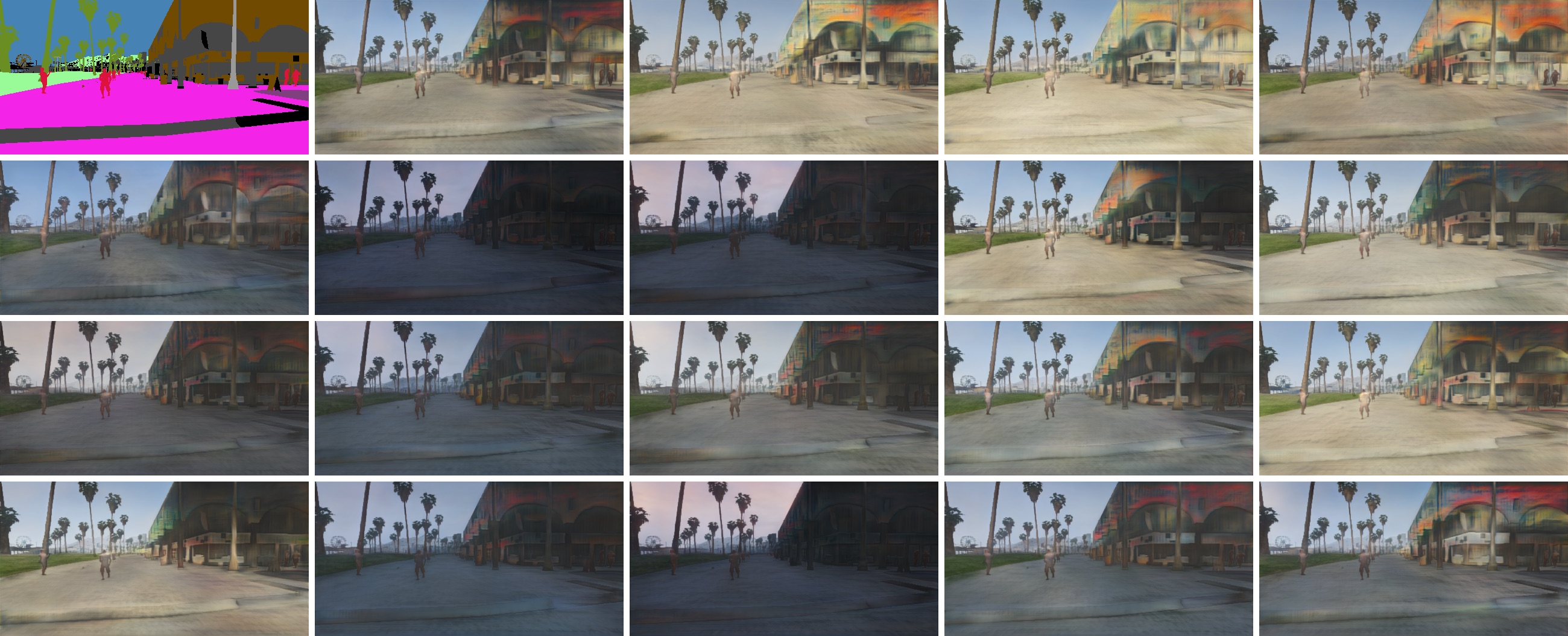}
    \caption{}
    \end{subfigure}
    \begin{subfigure}{\textwidth}
    \includegraphics[width=\linewidth]{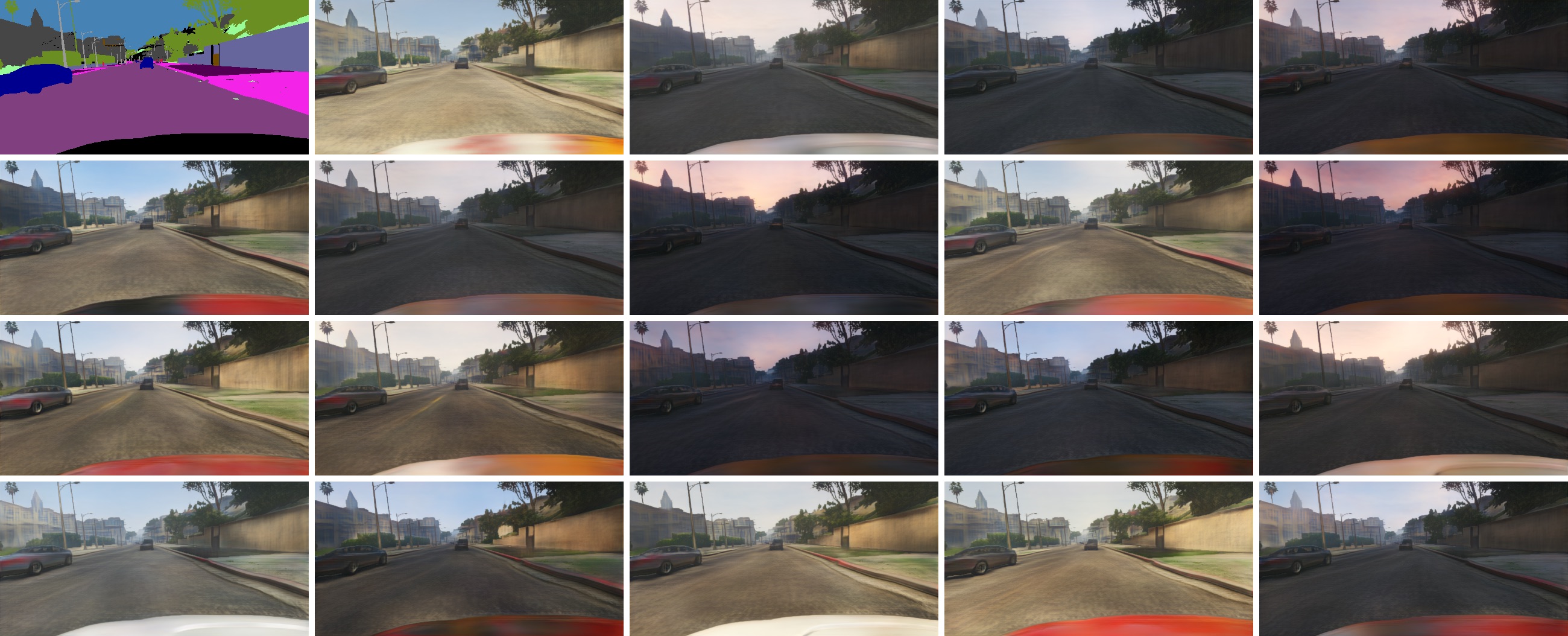}
    \caption{}
    \end{subfigure}
    \caption{Samples generated by our model. The image at the top-left corner is the input semantic layout and the other 19 images are samples generated by our model conditioned on the same semantic layout.}
    \label{fig:eg20_more2}
\end{figure*}

\end{appendices}

\end{document}